%% file: main.tex
\documentclass{article} 
\usepackage{iclr2026_conference,times}
\iclrfinalcopy

\usepackage{hyperref}
\usepackage{url}
\usepackage{ulem}

\input{utils/includes}

\input{utils/general_utils}
\input{utils/math_utils}

\input{utils/names}

\everydisplay{\small}

\title{{\ours}: Improving Reasoning Models via Cyclical Reflection Token Scheduling}

\author{Chongyu Fan$^{\dag}$ ~~Yihua Zhang$^{\dag}$ ~~Jinghan Jia$^{\dag}$ ~~Alfred Hero$^{\ddag}$ ~~Sijia Liu$^{\dag,\S}$\\
  $^\dag$Michigan State University\\
  $^\ddag$University of Michigan, Ann Arbor\\
  $^\S$IBM Research\\
}

\begin{document}

\maketitle

\input{sections/abstract}

\input{sections/introduction}

\input{sections/related_work}

\input{sections/preliminary_revision}

\input{sections/methodology_revision}

\input{sections/experiments}

\input{sections/conclusion}

\input{sections/acknowledgment}

\bibliography{refs/RA}
\bibliographystyle{iclr2026_conference}

\clearpage
\newpage
\appendix
\input{sections/appendix}

\end{document}

%% file: utils/includes.tex
\usepackage{wrapfig}

\usepackage{multirow,mathtools }

\usepackage{pifont}
\usepackage{color, colortbl}

\usepackage{blindtext}
\usepackage{lipsum}

\usepackage{multirow}
\usepackage{listings}

\usepackage{bbm}

\usepackage [english]{babel}
\usepackage[autostyle, english = american]{csquotes}

\usepackage{pifont}
\usepackage{url}
\usepackage[most]{tcolorbox}

\usepackage{lipsum}
\usepackage{soul}
\usepackage{xcolor}
\usepackage{wrapfig}
\usepackage{multirow,mathtools } 

\usepackage{adjustbox}
\MakeOuterQuote{"}

\usepackage{microtype}
\usepackage{graphicx}
\usepackage{subfigure}
\usepackage{booktabs} 

\usepackage{amsmath}

\usepackage[capitalize,noabbrev]{cleveref}

\usepackage{nicefrac}       

\usepackage{tablefootnote}

%% file: utils/general_utils.tex
\usepackage{pifont}

\usepackage{color, colortbl}
\definecolor{Gray}{gray}{0.93}
\definecolor{Orange}{rgb}{1,0.5,0}
\definecolor{DGray}{gray}{0.83}
\definecolor{LightCyan}{rgb}{0.88,1,1}

\definecolor{WarnREd}{rgb}{1,0.4,0.4}
\definecolor{WarnOrange}{rgb}{1,0.682,0.502}
\definecolor{WarnPink}{rgb}{0.9176, 0.7215, 0.7215}
\definecolor{GoodGreen}{rgb}{0.5019, 0.9215, 0.6039}

\usepackage[T1]{fontenc}

\definecolor{styleblue}{HTML}{504099}
\definecolor{mypurple}{HTML}{9391ff}

\definecolor{bluegray}{rgb}{0.4, 0.6, 0.8}
\definecolor{ceruleanblue}{rgb}{0.16, 0.32, 0.75}

\hypersetup{
colorlinks=true,
citecolor=ceruleanblue,
linkcolor=ceruleanblue,
urlcolor=black
}

\definecolor{Gray}{gray}{0.93}

\definecolor{mr}{RGB}{173, 216, 230}
\definecolor{best}{RGB}{251, 111, 111}

\usepackage[most]{tcolorbox}
\newtcolorbox{prompt}[2][]{colback=gray!5!white,colframe=gray!75!black,title=#2,#1}

\usepackage{makecell}


%% file: utils/math_utils.tex
\usepackage{amsmath,amsfonts,bm}

\def\eqref#1{(\ref{#1})}

\def\1{\bm{1}}

\DeclareMathAlphabet{\mathsfit}{\encodingdefault}{\sfdefault}{m}{sl}
\SetMathAlphabet{\mathsfit}{bold}{\encodingdefault}{\sfdefault}{bx}{n}



%% file: utils/names.tex
\newcommand{\ours}{CyclicReflex}

%% file: sections/abstract.tex
\begin{abstract}

Large reasoning models (LRMs), such as OpenAI's o1 and DeepSeek-R1, harness test-time scaling to perform multi-step reasoning for complex problem-solving. This reasoning process, executed before producing final answers, is often guided by special juncture tokens that prompt self-evaluative reflection. These transition markers and reflective cues are referred to as ``\textit{reflection tokens}'' (\textit{e.g.}, ``wait'', ``but'', ``alternatively''). In this work, we treat reflection tokens as a ``\textit{resource}'' and introduce the problem of \textit{resource allocation}, aimed at improving the test-time compute performance of LRMs by adaptively regulating the frequency and placement of reflection tokens. Through empirical analysis, we show that 
 both excessive and insufficient use of reflection tokens, referred to as over-reflection and under-reflection, can degrade model performance.
To better understand this trade-off, 
we draw an analogy between reflection token usage and learning rate scheduling in optimization. Building on this insight, We propose \textit{cyclical reflection token scheduling} (termed {\ours}), a training-free decoding strategy that dynamically modulates reflection token logits with a bidirectional, position-dependent triangular waveform, incurring no additional computation cost. Experiments on MATH500, AIME2024/2025, AMC2023, GPQA Diamond and LiveCodeBench demonstrate that {\ours} consistently improves performance across model sizes (1.5B–14B), outperforming standard decoding and recent approaches such as TIP (thought switching penalty) and S1. Codes are available at \,\url{https://github.com/OPTML-Group/CyclicReflex}.
\end{abstract}

%% file: sections/introduction.tex
\section{Introduction}
\label{sec: intro}

There has been a recent surge in the development of large reasoning models (LRMs), driven by the introduction of chain-of-thought (CoT) \citep{wei2022chain}. Notable examples include OpenAI’s o1 \citep{openai2024o1}, Qwen 2.5 \citep{yang2024qwen2}, DeepSeek-R1 \citep{guo2025deepseek}, and Kimi-1.5 \citep{team2025kimi}. These models perform multi-step reasoning by generating so-called reflection tokens, phrases such as \textit{“wait”, “but”, “alternatively”}, which signal hesitation, alternative exploration, or intermediate analysis. 
In parallel, test-time scaling \citep{snell2024scaling,liu2025can,sadhukhan2025kinetics} have emerged as a complementary approach for improving reasoning accuracy by expanding the breadth or depth of CoT traces during inference.

However, LRMs remain prone to reasoning failures due to mismanagement of reflection tokens, often resulting in either \textit{underthinking} or \textit{overthinking}. Underthinking occurs when the model fails to fully explore promising reasoning paths for complex problems, often terminating prematurely or switching strategies too soon \citep{wang2025thoughts,su2025between}. In contrast, overthinking arises when the model generates an excessive number of reflection tokens on simple problems, leading to unnecessary computational overhead \citep{chen2024not,kumar2025overthink}. These observations show that, as internal signals for deliberative reasoning, reflection tokens play a critical role in shaping answer quality. The emerging challenges further underscore the need for a principled mechanism to regulate reflection token usage during inference.

In this work, we introduce the concept of \textbf{resource allocation for LRMs}, treating reflection tokens as a valuable resource whose scheduling along the CoT trajectory (\textit{a.k.a.} reasoning trace) can be strategically designed to improve reasoning accuracy. The objective is to optimize the quantity and placement of reflection tokens, adapting dynamically to the reasoning schedule and the difficulty of the current problem.
For instance, some problems exhibit \textit{under-reflection}, where too few reflection tokens result in premature answer generation, while others suffer from \textit{over-reflection}, where excessive tokens stall progress by repeatedly looping on phrases like ``wait''. This raises a central question:

\begin{tcolorbox}[before skip=2mm, after skip=0.0cm, boxsep=0.0cm, middle=0.0cm, top=0.05cm, bottom=0.05cm, boxrule=0.6pt]
\begin{center}
     \textit{\textbf{(Q)} How can we achieve effective resource allocation in LRMs to mitigate both under-reflection and over-reflection?}
\end{center}
\end{tcolorbox} 
\vspace*{2mm}

To answer this, \textbf{we draw a conceptual analogy between reflection tokens in LRMs and learning rates in optimization}. Leveraging the {landscape of thoughts} \citep{zhou2025landscape}, we show that under-reflection mirrors the effect of an overly small learning rate, leading to convergence to suboptimal solutions, while over-reflection resembles a large learning rate that causes divergence. We briefly introduce our motivation and the underlying intuition below.

\noindent 
\textbf{Overview of motivation and rationale: From stepsize hedging to cyclical learning rates.}
The critical role of learning rates (also known as stepsizes) in shaping optimization dynamics has been extensively studied~\citep{nesterov1983method, allen2014linear, bubeck2015geometric}. A recent theoretical advancement, the \textit{silver stepsize schedule}~\citep{altschuler2024acceleration, altschuler2025acceleration}, demonstrates that replacing a constant learning rate with an approximately periodic, \textit{hedging-style} schedule can \textit{provably accelerate convergence} in gradient descent. This approach is known as \textit{stepsize hedging} as it alternates strategically between large and small stepsizes, balancing rapid (but potentially unstable) exploration with slower, more stable convergence. A similar stepsize hedging idea has been applied to deep model training through \textit{cyclical learning rate schedule}~\citep{smith2017cyclical}, which alternates between large and small learning rates in a triangular waveform. This strategy not only accelerates convergence but also enhances generalization, often eliminating the need for extensive hyperparameter tuning.

\begin{wrapfigure}{r}{0.6\textwidth}
\centering
\vspace{-4mm}
\includegraphics[width=1.0\linewidth]{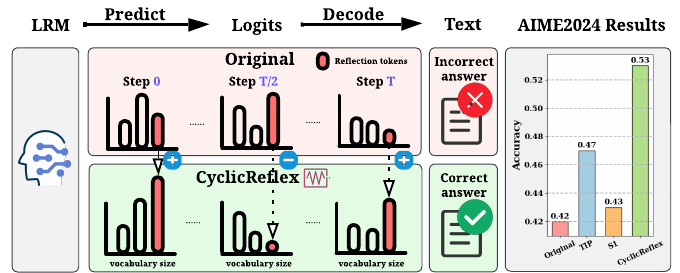}
\caption{\small{
Schematic overview of our proposed method ({\ours}). The rightmost subfigure presents a comparison of final answer accuracy between {\ours}, the original LRM (DeepSeek-R1-Distill-Llama-8B), and decoding variants using TIP~\citep{wang2025thoughts} and S1~\citep{muennighoff2025s1}.}
}
\vspace{-3mm}
\label{fig: overview}
\end{wrapfigure}

Motivated by the principle of stepsize hedging \citep{altschuler2024acceleration, altschuler2025acceleration} and the demonstrated effectiveness of cyclical learning rates in deep learning~\citep{smith2017cyclical}, we propose \textbf{{\ours}}, a training-free decoding strategy that dynamically modulates the logits of reflection tokens using a position-dependent, periodic triangular waveform (see \textbf{Fig.\,\ref{fig: overview}}). Just as cyclical learning rates alternate between aggressive and conservative updates to balance exploration and convergence, {\ours} cyclically adjusts the sampling likelihood of reflection tokens to regulate the depth and stability of the reasoning process.
Unlike conventional decoding methods, {\ours} is \textit{bidirectional}, capable of both promoting and suppressing reflection token usage depending on the stage of generation. This flexibility enables {\ours} to address both under-reflection (insufficient reasoning) and over-reflection (excessive, redundant reasoning), offering a principled mechanism for reasoning modulation inspired by optimization dynamics.

\noindent \textbf{Contributions.}
We summarize our contributions below.

$\bullet$  We introduce and formalize the  problem of resource allocation in LRMs by treating reflection tokens as a computational resource, motivated by the dual challenges of under-reflection and over-reflection in reasoning generation.

$\bullet$  We draw a conceptual analogy between reflection token scheduling and learning rate scheduling in optimization, and validate it through the landscape of thoughts. Guided by that, we propose {\ours}, a test-time decoding strategy that cyclically modulates reflection token logits to dynamically balance reflection during generation.

$\bullet$  We conduct comprehensive experiments across six reasoning benchmarks and multiple model scales (1.5B–8B), demonstrating that {\ours} consistently improves both final-answer accuracy and self-correction capability, outperforming recent methods such as TIP \citep{wang2025thoughts} and S1 \citep{muennighoff2025s1}. Moreover, {\ours} integrates seamlessly with other test-time scaling techniques, yielding additional performance gains.

%% file: sections/related_work.tex
\section{Related Work}
\label{sec: related_work}

\noindent \textbf{LRMs and CoT.}
CoT \citep{wei2022chain} enables LRMs to solve complex tasks through intermediate reasoning steps before reaching a final answer. This technique underpins many recent LRMs, including OpenAI’s o1 \citep{openai2024o1}, Qwen 2.5 \citep{yang2024qwen2}, DeepSeek-R1 \citep{guo2025deepseek}, and Kimi-1.5 \citep{team2025kimi}, which often employ reinforcement learning to further improve their reasoning performance. \citet{guo2025deepseek} show that even smaller models benefit substantially from fine-tuning with CoT-style data. A hallmark of CoT reasoning in these models is the emergence of reflection tokens (words like ``wait'' or ``but'') that signal deliberation or self-correction, marking a shift from fast to slow thinking \citep{kumar2025llm,li2025system}. In this paper, we show that the reasoning performance of LRMs can be enhanced by applying cyclical logits manipulation to reflection tokens.

\noindent \textbf{Efficient reasoning.}
Despite their impressive capabilities, LRMs often exhibit reasoning inefficiencies. Overthinking arises when the model generates unnecessarily long reasoning traces, leading to inflated outputs and increased computational cost \citep{chen2024not,kumar2025overthink}. In contrast, underthinking occurs when the model halts reasoning too early, failing to adequately explore promising solution paths \citep{wang2025thoughts,su2025between}.
Therefore, ensuring both the efficacy (\textit{i.e.}, answer accuracy) and efficiency (\textit{i.e.}, generation length) of reasoning is crucial.  Building on this line of research, some approaches modify model behavior through post-training interventions. For instance, \citet{luo2025o1,aggarwal2025l1,hou2025thinkprune}
use fine-tuning or reinforcement learning to explicitly control reasoning length. 
There also exist works that adopt training-free strategies. 
\citet{yang2025speculative} propose guiding smaller models with larger ones at inference; \citet{wang2025thoughts} penalize reflection token logits to reduce over-reflection; And \citet{yang2025dynamic,muennighoff2025s1} develop early-exit mechanisms for efficient decoding. Our method also falls into the training-free category but differs in its dynamic to adaptively address both under- and over-reflection without model modification.

\noindent\textbf{Test-time scaling.} 
A growing body of work enhances LRM reasoning via test-time scaling. Basic strategies include manually inserting reflection tokens (\textit{e.g.}, ``wait”, ``but'') to prompt deeper thinking \citep{muennighoff2025s1,jin2025well,chen2025iterative}. More methods such as Best-of-N generation and self-consistency sampling \citep{wang2022self,irvine2023rewarding,brown2024large,kang2025scalable,wang2025every} aim to select the most promising answer among multiple candidates, often guided by reward models. Structured decoding approaches, such as beam search \citep{feng2023alphazero}, tree-of-thought (ToT) \citep{yao2023tree}, and Monte Carlo tree search (MCTS) \citep{zhou2023language}, further improve answer quality by enabling the model to reason over multiple candidate paths. In pursuit of controlled reasoning, \citet{wu2025effectively} propose thinking intervention, which selectively inserts or edits specific thinking tokens during generation to tailor LRM behavior for downstream tasks. Recent analyses \citep{snell2024scaling,liu2025can,chen2025seal,zhang2025alphaone,ghosal2025does} also highlight that the effectiveness of test-time scaling varies with problem difficulty, motivating strategies that adapt to instance complexity. Our method provides the adaptive test-time compute by dynamically adjusting the influence of reflection tokens throughout the reasoning trajectory. We show that it consistently improves accuracy across difficulty levels and can be integrated seamlessly with other test-time strategies like Best-of-N and beam search.

%% file: sections/preliminary_revision.tex
\section{Resource Allocation in Reasoning Models}
\label{sec: preliminary}

In this section, we begin by introducing preliminaries on LRMs, including their chain-of-thought trajectories (\textit{i.e.}, reasoning traces) and the use of reflection tokens. We then motivate the problem of resource allocation over reflection tokens through an existing technique: \underline{t}hought sw\underline{i}tching \underline{p}enalty (TIP) test-time compute strategy. This warm-up study illustrates the critical influence and sensitivity of reflection tokens on the final answers produced by LRMs.

\noindent \textbf{Preliminaries on LRMs, reasoning traces, and reflection tokens.}
Unlike conventional LLMs, LRMs can incorporate an explicit \textit{thinking stage} before arriving at a {final answer} \citep{li2025system,kumar2025llm,chen2025towards}. This thinking stage is typically realized through a CoT trajectory composed of multiple reasoning steps.

\begin{wraptable}{r}{0.45\textwidth}
\centering
\vspace*{-8mm} 
\caption{\small{An example from MATH500 answered by DeepSeek-R1-Distill-Qwen-7B. Reflection tokens are highlighted in \textcolor{red}{red}.}}
\label{tab: lrm_example}
\resizebox{1.0\linewidth}{!}{
\begin{tabular}{>{\centering\arraybackslash}m{0.5cm}|>{\raggedright\arraybackslash}m{7.8cm}}
\toprule[1pt]
\midrule
 & \multicolumn{1}{c}{\textbf{Generation ($\mathbf r$ and $\mathbf y$) vs. input $\mathbf x$}} \\
\midrule
$\mathbf{x}$ & Compute $\arcsin \left( -\frac{1}{2} \right).$ Express your answer in radians. \\
\midrule
$\mathbf{r}$ & \textbf{<think>}Okay, so I need to compute arcsin of -1/2...... \textcolor{red}{But} I should make sure I'm thinking about the right range for arcsin. I think it's between $-\pi/2$ and $\pi/2$, right? So it's the principal value...... I know that sine is negative in the third and fourth quadrants, \textcolor{red}{but} since arcsin only gives me angles in $[-\pi/2, \pi/2]$, that's the fourth quadrant...... \textcolor{red}{Wait}, let me double-check. If I take $\sin(-\pi/6)$, that's the same as $-\sin(\pi/6)$, which is -1/2. So the answer should be $-\pi/6$ radians.\textbf{</think>} \\
\midrule
$\mathbf{y}$ & The arcsin of -1/2 is $-\pi/6$ radians. \\
\midrule
\bottomrule[1pt]
\end{tabular}
}
\vspace*{-4mm} 
\end{wraptable}

To be concrete, let $\mathbf{x}$ denote the input question, $\mathbf{r}$ the reasoning trace, and $\mathbf{y}$ the final answer. The reasoning trace can be written as $\mathbf{r} = [\mathbf{r}_1, \mathbf{r}_2, \ldots, \mathbf{r}_T]$, a sequence of $T$ intermediate reasoning steps (or thoughts) that serve as a precondition for producing the final answer $\mathbf{y}$. These thoughts are often prefixed by lexical cues such as ``\textit{wait}'', ``\textit{but}'', and ``\textit{alternatively}''.
In this work, we collectively refer to such words as \textbf{reflection tokens}, as they signal the LRM’s introspective or deliberative thinking during generation. These tokens typically signal hesitation, reconsideration or alternative exploration within the reasoning trace, serving as implicit markers of the model’s engagement in step-by-step reasoning.
For ease of understanding, \textbf{Table\,\ref{tab: lrm_example}} illustrates an example where an LRM tackles a math problem ($\mathbf{x}$), along with its generated reasoning trace ($\mathbf{r}$), final answer ($\mathbf{y}$), and the highlighted reflection tokens embedded within the reasoning process.

\noindent \textbf{Motivation: Influence of reflection tokens on final answer accuracy.}
Next, we present a \textit{warm-up} study to illustrate the critical role of reflection tokens in determining final answer quality, particularly across varying levels of problem difficulty.
Prior work has identified several common issues in LRM, including ``underthinking'' \citep{wang2025thoughts,su2025between}--where the model prematurely abandons promising lines of thought, resulting in insufficient reasoning depth--and ``overthinking'' \citep{muennighoff2025s1,chen2024not,kumar2025overthink}--where excessive, unnecessary reasoning steps obscure or derail the correct final answer. These studies have observed that intervening in the reasoning trace with reflection tokens can help guide the model’s thought process \citep{wu2025effectively}.
A common approach to controlling and integrating the effect of reflection tokens in LRM generation is to modify the decoding strategy to account for their occurrence. One such method is TIP \citep{wang2025thoughts}, which was proposed to discourage the generation of reflection tokens and thereby penalize frequent thought switches during the reasoning trace. 

Given the set of reflection tokens $\hat{V}$, TIP introduces a logit penalty ($\alpha$) to the predicted score $z_{t,v}$ when generating a reflection token $v \in \hat{V}$ at reasoning step $t$, yielding the updated logit
\begin{align}
\mathrm{TIP}(\alpha): \quad \hat{z}_{t,v} = \left \{ 
\begin{array}{ll}
z_{t,v} + \alpha, & \text{if } v \in \hat{V} \text{ and } t < T_0 \\
z_{t,v}, & \text{otherwise}
\end{array}
\right.
\label{eq:TIP}
\end{align}
where $\alpha$ controls the strength of the logit intervention, and $T_0$ specifies the time window over which the adjustment is applied. For clarity,  TIP($\alpha$) denotes the TIP-based decoding strategy parameterized by $\alpha$. It is worth noting that \textit{TIP sets $\alpha \leq 0$ to penalize frequent thought switches}. 

\begin{figure*}[htbp]
\centering
\begin{tabular}{ccc}
\hspace*{-3mm}
\includegraphics[width=0.225\textwidth]{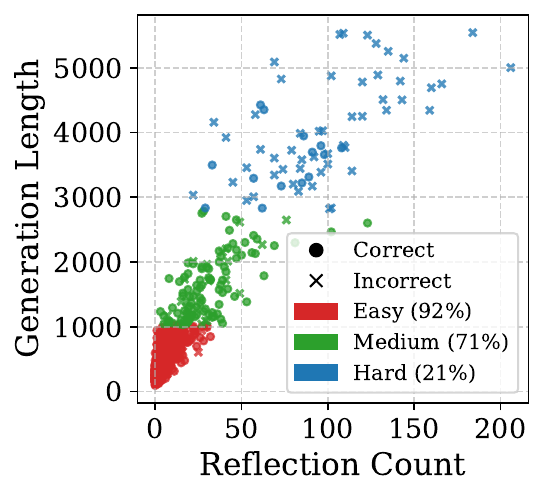} 
&
\hspace*{-5mm}
\includegraphics[width=0.23\textwidth]{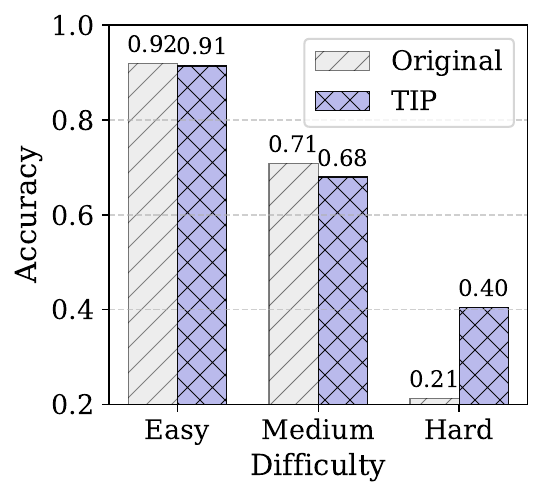}
&
\hspace*{-5mm}
\includegraphics[width=0.53\textwidth]{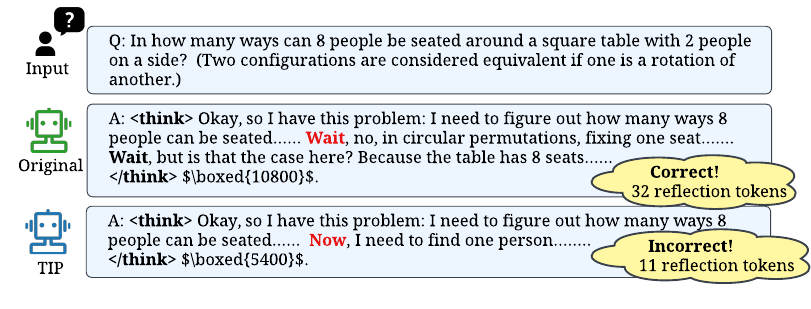}
\\
\small{(a) Difficulty groups} 
&  
\small{(b) Accuracy}
&  
\small{(c) Generation examples}
\end{tabular}
\vspace{-3mm}
\caption{\small{(a) Answers from DeepSeek-R1-Distill-Qwen-7B on MATH500 clustered into Easy, Medium, and Hard using K-means over reflection word count and generation length. Each point represents one answer. (b) Accuracy of original decoding and TIP across difficulty levels. (c) Generation examples of original decoding and TIP for a problem from the Medium category.}
}
\vspace{-2mm}
\label{fig: tip_motivation}
\end{figure*}

Despite the heuristic nature of choosing $\alpha$ and $T_0$, the TIP-based decoding strategy \citep{wang2025thoughts} provides useful motivation for studying resource allocation over reflection tokens and its impact on reasoning effectiveness (\textit{e.g.}, on the MATH500) across \textit{different problem difficulty levels}. As shown in \textbf{Fig.\,\ref{fig: tip_motivation}(a)}, we categorize MATH500 problems into three difficulty levels---Easy, Medium, and Hard---based on generation length and reflection token count, with the final-answer accuracies  92\%, 71\%, and 21\%, respectively. We observe that for more challenging problems, LRMs tend to produce longer reasoning trajectories and more reflection tokens, indicating deeper engagement in problem-solving. \textbf{Fig.\,\ref{fig: tip_motivation}(b)} next compares the accuracy of the original decoding strategy and TIP across the difficulty groups in Fig.\,\ref{fig: tip_motivation}(a). As shown, TIP improves accuracy on Hard problems \textit{but reduces accuracy on Easy and Medium problems}. This suggests that TIP’s constant logit manipulation strategy (agnostic to the reasoning step $t$) does not yield \textit{optimal} reasoning control.
Furthermore, \textbf{Fig.\,\ref{fig:  tip_motivation}(c)} shows a Medium-level example comparing original decoding with TIP. The first divergence in reasoning is highlighted in red. Under original decoding, the model introduces a transitional reflection (``Wait, no, in circular permutations, fixing one seat...''), generating 32 reflection tokens before arriving at the correct answer. In contrast, with a thought-switching penalty $\alpha < 0$ in \eqref{eq:TIP}, TIP reshapes the trace (``Now, I need to find one person...''), producing only 11 reflection tokens and yielding an incorrect answer. This shows that TIP provides only one-directional reflection control (penalizing reflection token logits). Hence, a bi-directional, dynamically adaptive (non-constant) reflection token allocation strategy is needed.

\noindent \textbf{Problem of interest: Resource allocation over reflection tokens.}
Reflection tokens have a significant impact on the reasoning capability of LRMs. Therefore, if we view reflection tokens as a ``resource'' in LRM reasoning generation, then determining their schedule, including the number of occurrences and their positions, naturally gives rise to the \textit{problem of resource allocation} for LRMs. 

TIP offers a simple solution by applying a constant logit penalty to reflection token generation. However, this approach is \textit{static} and therefore fails to account for both the number and placement of reflection tokens, which are dynamically determined during reasoning trace generation.
As shown in Fig.\,\ref{fig: tip_motivation}(b), TIP does not consistently improve performance across all difficulty levels.
This leaves open the question of how to schedule reflection token generation along the reasoning trajectory, that is, how to allocate these ``resources'' effectively over time while accounting for problem difficulty. These underscore the need for more adaptive and fine-grained strategies to control reflection token usage in order to address the resource allocation more effectively.

%% file: sections/methodology_revision.tex
\section{Reflection Token Scheduling  as Learning Rate Scheduling in Optimization}
\label{sec: method}

In this section, we draw a conceptual analogy between reflection token scheduling and learning rate scheduling in optimization, aimed at deepening our understanding of reflection tokens in reasoning and enabling more effective resource allocation.
And we propose a new decoding strategy: cyclical reflection token scheduling (\textbf{\ours}).

\noindent \textbf{Reflection tokens in the thought landscape vs. learning rates in the optimization landscape.}
The role of reflection tokens in reasoning closely mirrors that of learning rates in optimization. In the ``thought landscape'', a model initiates by interpreting a question and leverages reflection tokens to modulate its reasoning trajectory: exploring, reconsidering, and refining intermediate steps before reaching a final answer. Likewise, in the optimization landscape, an optimizer begins from a random initialization and relies on the learning rate to control the step size of the variable updates, gradually converging toward an optimal solution. In both cases, a well-tuned control mechanism, reflection tokens in reasoning or learning rates in optimization, is essential for accurate solution convergence.

Additionally, in optimization, an improperly tuned learning rate, either too small or too large, can hinder convergence, causing the optimizer to either stagnate or diverge. This challenge in scheduling the learning rate maps onto the difficulty of scheduling reflection tokens in reasoning, manifesting as under-reflection and over-reflection.

(\textit{Under-reflection}) When the model generates too few reflection tokens, it often terminates the reasoning process prematurely, resulting in a final answer that lacks sufficient deliberation. This is analogous to optimization with a learning rate that is too small, where the model converges too early and becomes trapped in a suboptimal local minimum.

(\textit{Over-reflection}) Conversely, generating too many reflection tokens can prevent the model from concluding its reasoning, causing it to loop or stall, \textit{e.g.}, repeatedly producing phrases like ``wait'' without reaching a solution. This resembles optimization with an overly large learning rate, which leads to instability and divergence rather than convergence.

To validate the analogy between reflection tokens and learning rates (too small learning rate vs. under-reflection, and too large learning rate vs. over-reflection), we utilize the interpretability tool introduced in \citep{zhou2025landscape} to visualize the \textit{landscape of thoughts}. This tool projects reasoning step $\mathbf{r}_i$ into a two-dimensional visual space based on the measured ``distance'' between each step $\mathbf{r}_i$ and the final answer $\mathbf{y}$, providing an interpretable view of the model’s reasoning dynamics. The distance metric captures the model’s uncertainty by taking the inverse of the probability of generating the answer $\mathbf{y}$ conditioned on the reasoning step $\mathbf{r}_i$, normalized by the length of $\mathbf{y}$:
\begin{align}
d(\mathbf{r}_i, \mathbf{y}) = p_{\mathrm{LRM}}(\mathbf{y} \mid \mathbf{r}_i)^{-1/|\mathbf{y}|},
\end{align}
where $p_{\mathrm{LRM}}$ denotes the prediction probability assigned by the LRM to the answer $\mathbf{y}$ given the reasoning step $\mathbf{r}_i$, and $|\mathbf{y}|$ denotes the length of $\mathbf{y}$.

\begin{figure}[htb]
    \centering
    \includegraphics[width=0.8\textwidth]{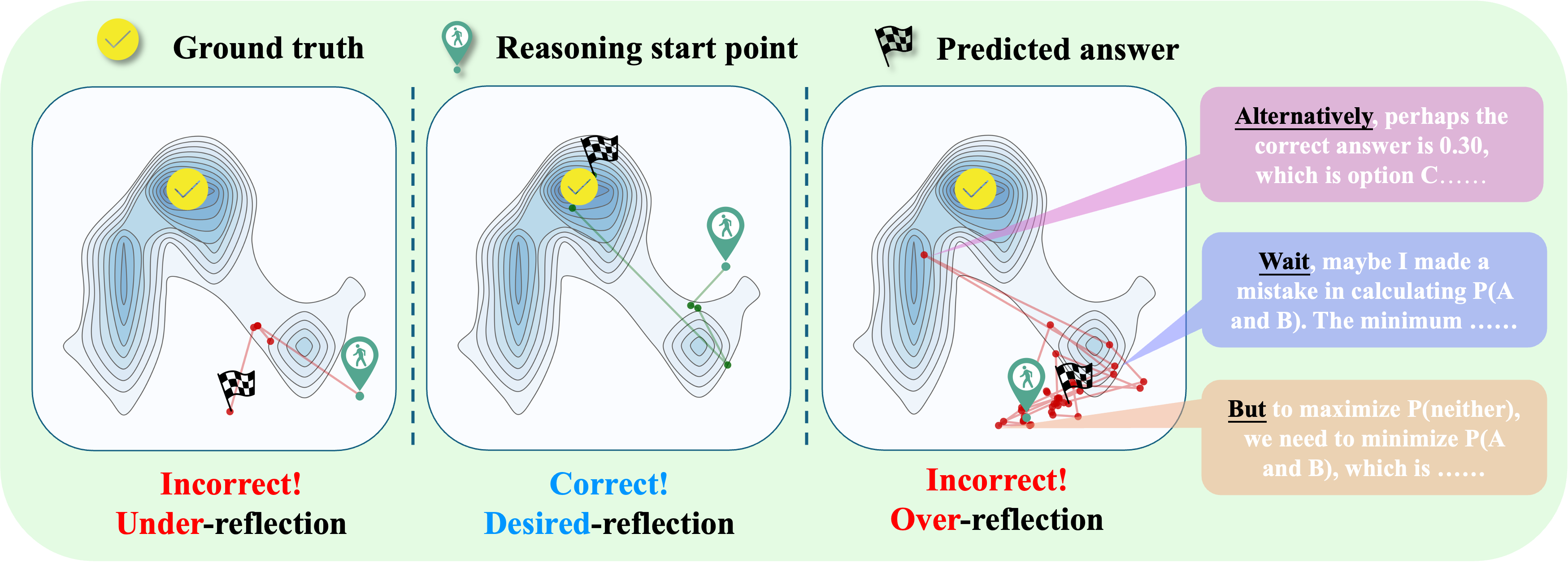}
    \vspace{-2mm}
    \caption{\small{
    Examples of landscape of thought for under-reflection, desired-reflection, and over-reflection, generated by DeepSeek-R1-Distill-Qwen-7B with the original decoding strategy. Each point represents a reasoning step and is connected in the order of generation. Darker regions indicate steps with higher semantic alignment to the correct answer.
    }}
    \label{fig: lot}
\end{figure}

\textbf{Fig.\,\ref{fig: lot}} presents a visualized reasoning trajectory from the initial thought to the final answer under original decoding strategy, across three different scenarios: (i) under-reflection, where too few reflection tokens lead to a reasoning trace that is too short and results in an incorrect answer; (ii) desired reflection, which yields a well-structured reasoning trace and a correct answer; and (iii) over-reflection, where excessive reflection tokens cause an overly long and off-track reasoning trace, also resulting in an incorrect answer. In the landscape, darker regions represent intermediate reasoning steps that are semantically closer to the correct answer. That is,   color intensity reflects the relative correctness of each thought along the trajectory.

As we can see, the thought landscape under under-reflection is too conservative to drive the reasoning process away from the starting point, ultimately failing to converge to the correct final answer. In contrast, over-reflection could enable the model to reach semantically promising regions of the landscape, for example, a step like ``\textit{Alternatively, perhaps the correct answer is ...}'', which is far away from the thinking start point and located in the darker region.
However, much like an excessively large learning rate that fails to properly control the optimization process, this leads the model to quickly pass through the desirable state without settling there, ultimately leading to an incorrect answer region. Moreover, we find that reflection tokens are responsible for the \textit{sharp turns} in the reasoning trajectory. By examining the sharply turning steps in over-reflection, we observe that they are consistently initiated by reflection tokens.

\noindent \textbf{\ours: Cyclical logits manipulation for reflection token scheduling.}
Although reflection tokens are crucial for guiding multi-step reasoning, balancing their use remains challenging.
The need for dynamic modulation of reflection tokens closely mirrors the challenge of learning rate scheduling in optimization. As introduced in Sec.\,\ref{sec: intro}, the convergence of gradient descent can be \textit{provably accelerated} by adopting the \textit{silver stepsize schedule}, which follows the principle of \textit{stepsize hedging}~\citep{altschuler2024acceleration, altschuler2025acceleration}. The key algorithmic insight is to hedge between two individually suboptimal strategies, small and large stepsizes, since the failure modes of one are often mitigated by the strengths of the other.
In deep model training, cyclical learning rates \citep{smith2017cyclical} exemplify this principle in practice. Rather than using a fixed learning rate, they employ a triangular waveform to periodically alternate between large and small step sizes. This schedule allows the optimizer to balance global exploration (enabled by large steps) with local convergence stability (provided by smaller steps), thereby yielding a form of stepsize hedging.

\begin{wrapfigure}{r}{0.3\textwidth}
\centering
 \vspace*{-6mm}
\includegraphics[width=1.0\linewidth]{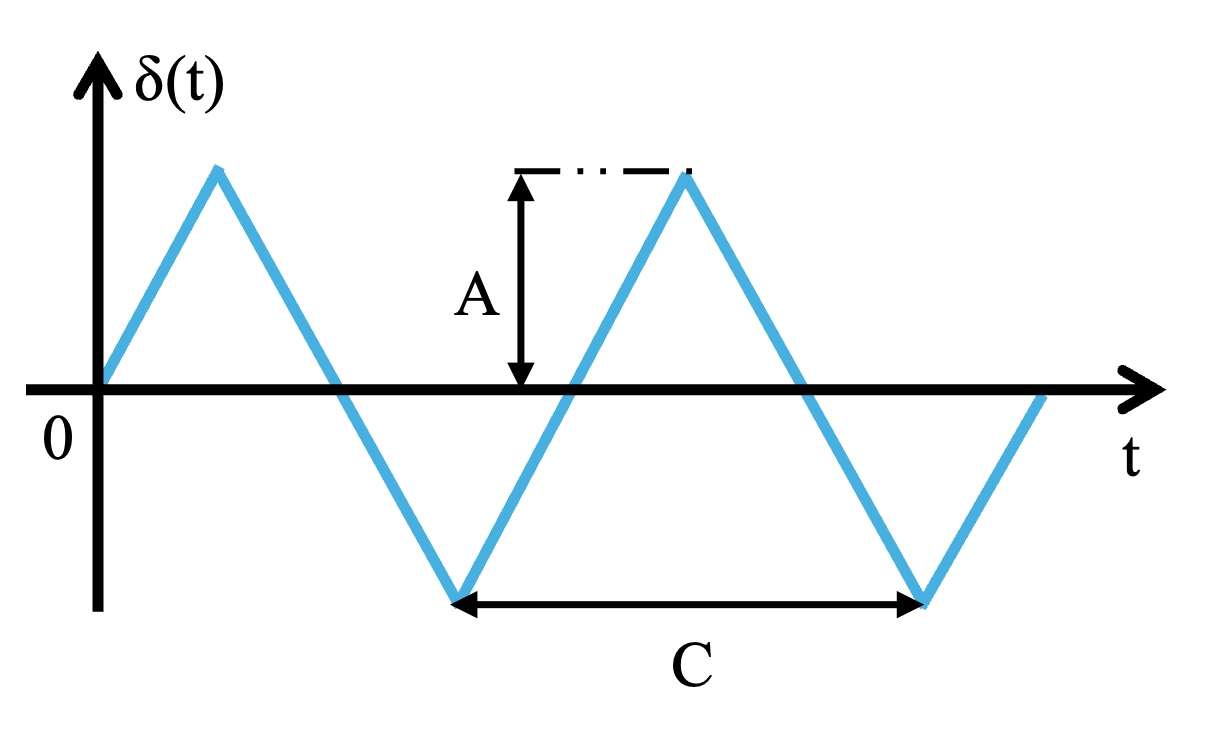}
\vspace*{-6mm}
\caption{\small{
Illustration of {\ours} \eqref{eq:cyclic}, where $t$ denotes the token position and $\delta(t)$ the logit adjustment on reflection tokens, oscillating between $-A$ and $A$ with amplitude $A$ and period $C$.
}}
\vspace*{-4mm}
\label{fig: cyclic-plot}
\end{wrapfigure}

Inspired by cyclical learning rates, we introduce \textbf{\ours}. As depicted in \textbf{Fig.\,\ref{fig: cyclic-plot}}, we apply a periodic triangular waveform to modulate the logits of reflection tokens during generation. 
The waveform is governed by two parameters: the amplitude $A$, which controls the strength of the logit adjustment, and the period 
$C$, which determines the oscillation frequency. This logits manipulation evolves over time and is both \textit{position-dependent} (varying with each decoding step) and \textit{bidirectional} (allowing for the dynamic promotion or suppression of reflection token sampling based on the current stage of the reasoning process).
More concretely, {\ours} can be cast as the following logits manipulation as the function of the reasoning step $t$
\begin{align}
\text{\ours}:~ 
\hat{z}_{t,v} &=
\left\{
\begin{array}{ll}
z_{t,v} + \delta(t) & \text{if } v \in \hat{V}, \\ 
z_{t,v}            & \text{otherwise,}
\end{array}
\right. &
\delta(t) &=
A \left| 
4 \cdot \frac{(t - \frac{C}{4}) \bmod C}{C} - 2
\right| - A
\label{eq:cyclic}
\end{align}
where recall that the amplitude $A$ and the period $C$ have been previously defined as shown in Fig.\,\ref{fig: cyclic-plot}, $\bmod$ is the modulo operation, $|\cdot|$ is the absolute value operation, and the other notations follow \eqref{eq:TIP}. In \eqref{eq:cyclic}, $(t - \tfrac{C}{4}) \bmod C$ gives the current thought position within the cycle, and it is straightforward to validate that $\delta(C/4) = A$ and $\delta(3C/4) = - A$.

As shown in Fig.\,\ref{fig: cyclic-plot}, \textit{{\ours} adopts a representative form of hedging schedule}: the increasing phase of the reflection logit adjustment $\delta(t)$ promotes exploration by encouraging the model to transition away from its current line of thought, while the decreasing phase fosters convergence by stabilizing the reasoning process, guiding the model toward producing a coherent final answer.

Compared to TIP \eqref{eq:TIP}, which applies a fixed unidirectional penalty, {\ours} adaptively modulates reflection token logits \sout{with} at no additional computation cost, offering finer control over reasoning. This unified mechanism balances under- and over-reflection, yielding more robust and flexible behavior that adapts to the model’s evolving thought process. In \textbf{Fig.\,\ref{fig: hyper}} of \textbf{Appendix\,\ref{appendix: add_exp}}, we summarize the sensitivity of {\ours} to its hyperparameters. The period $C$ has a larger impact than the amplitude $A$.

%% file: sections/experiments.tex
\section{Experiments}
\label{sec: exp}
\vspace*{-2mm}
\subsection{Experimental Setup}
\vspace*{-2mm}

\noindent \textbf{Data-model setups.} To evaluate the effectiveness of {\ours}, we consider both \textit{math} and \textit{non-math} benchmarks. The math datasets include \textbf{MATH500} \citep{lightman2023let} with 500 multi-step problems, \textbf{AIME2024/2025} \citep{aime} with 30 challenging problems each year, and \textbf{AMC2023} \citep{AMC2023} covering diverse competition topics. The non-math datasets include \textbf{GPQA Diamond} \citep{rein2024gpqa}, a challenging subset of multiple-choice science questions in biology, chemistry, and physics, and \textbf{LiveCodeBench} \citep{jain2024livecodebench}, a coding benchmark from LeetCode, AtCoder, and Codeforces that evaluates code generation, repair, and execution. We evaluate our method using {DeepSeek-R1-Distilled-Qwen} (1.5B, 7B) \citep{guo2025deepseek} and {DeepSeek-R1-Distilled-Llama-8B} to assess different backbones. Additionally, we demonstrate generalization on Qwen3 (4B, 8B, 14B) \citep{yang2025qwen3} in \textbf{Table\,\ref{tab: qwen}} of \textbf{Appendix\,\ref{appendix: add_exp}}.

\noindent \textbf{Baseline and evaluation.}
Our method ({\ours}) is compared against two primary baselines: \textbf{TIP} \citep{wang2025thoughts}, \textbf{S1} \citep{muennighoff2025s1}. In addition, we assess the compatibility of {\ours} with external test-time scaling techniques, including \textbf{Best-of-N} \citep{irvine2023rewarding,brown2024large} and \textbf{Beam Search} \citep{feng2023alphazero}, using RLHFlow-PRM-Deepseek-8B as the preference reward model (PRM) for scoring \citep{dong2024rlhf}. We use \textbf{accuracy} and \textbf{generation length} as our primary evaluation metrics. Accuracy is obtained by rule-based extraction of the final answer against the ground truth, while generation length is the total word count of the response. More implementation details are provided in Appendix\,\ref{appendix: exp_setup}. 

\input{tab/model_comparison}

\subsection{Experiment Results} 

\noindent \textbf{Overall Performance of {\ours} on the MATH Task.} In \textbf{Table\,\ref{tab: model_comparison}}, we show the effectiveness of 
\begin{wrapfigure}{r}{0.45\linewidth}
    \centering
    \hspace{6mm}
    \includegraphics[width=0.4\textwidth]{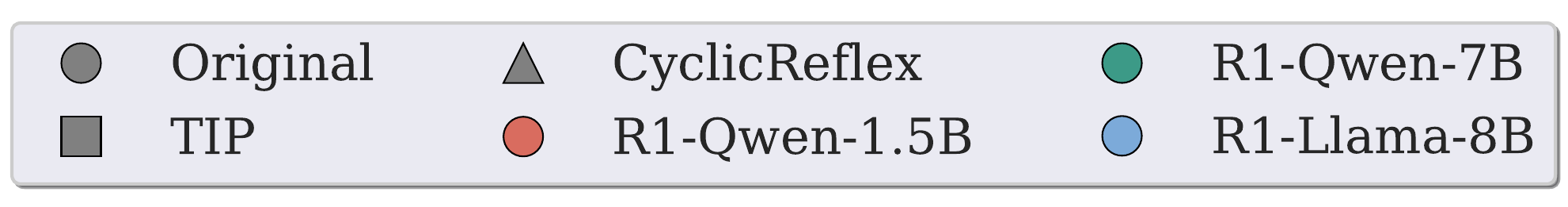}\\
    \begin{tabular}{cc}
        \includegraphics[width=0.22\textwidth]{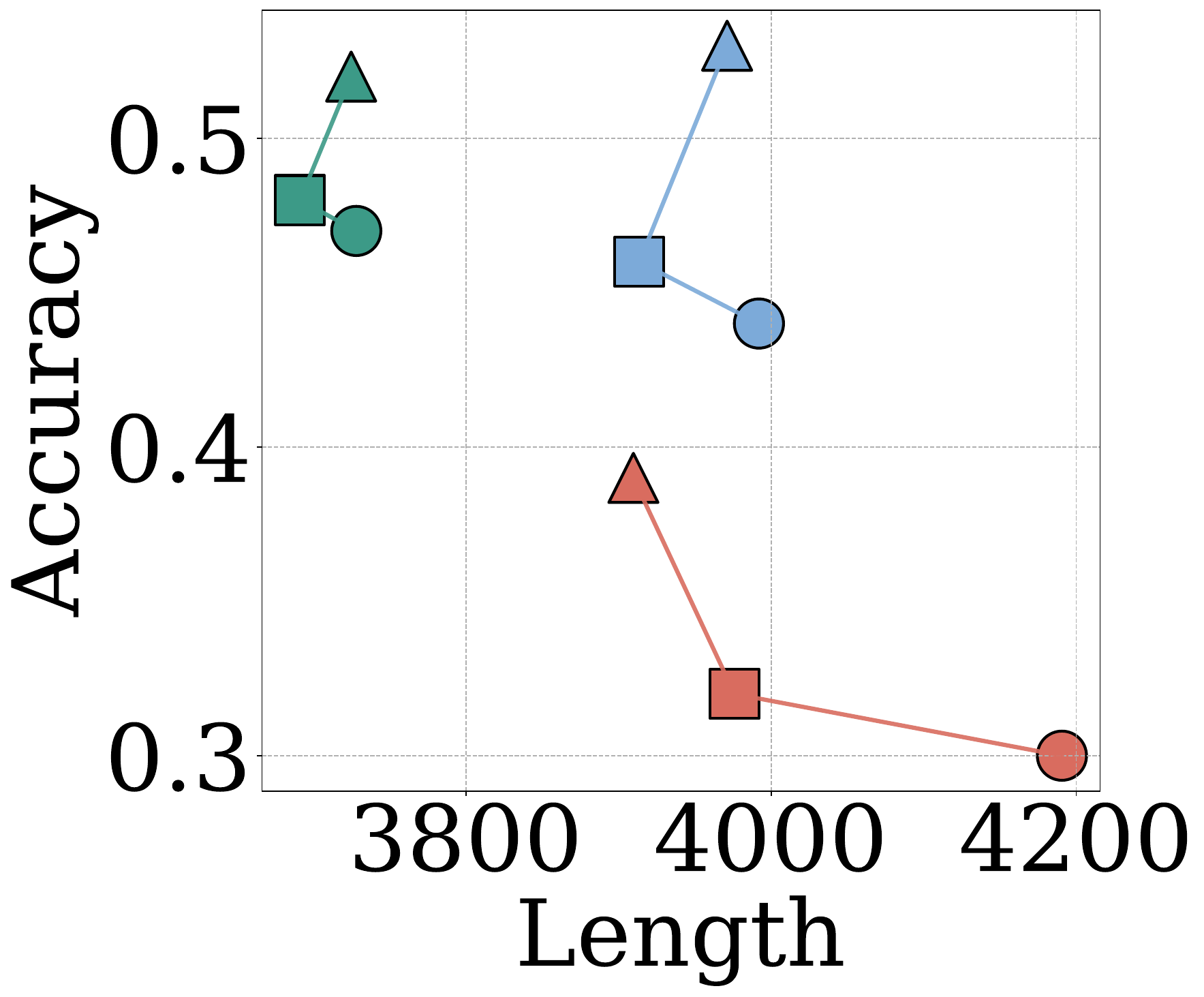} &
        \hspace{-5 mm}
        \includegraphics[width=0.22\textwidth]{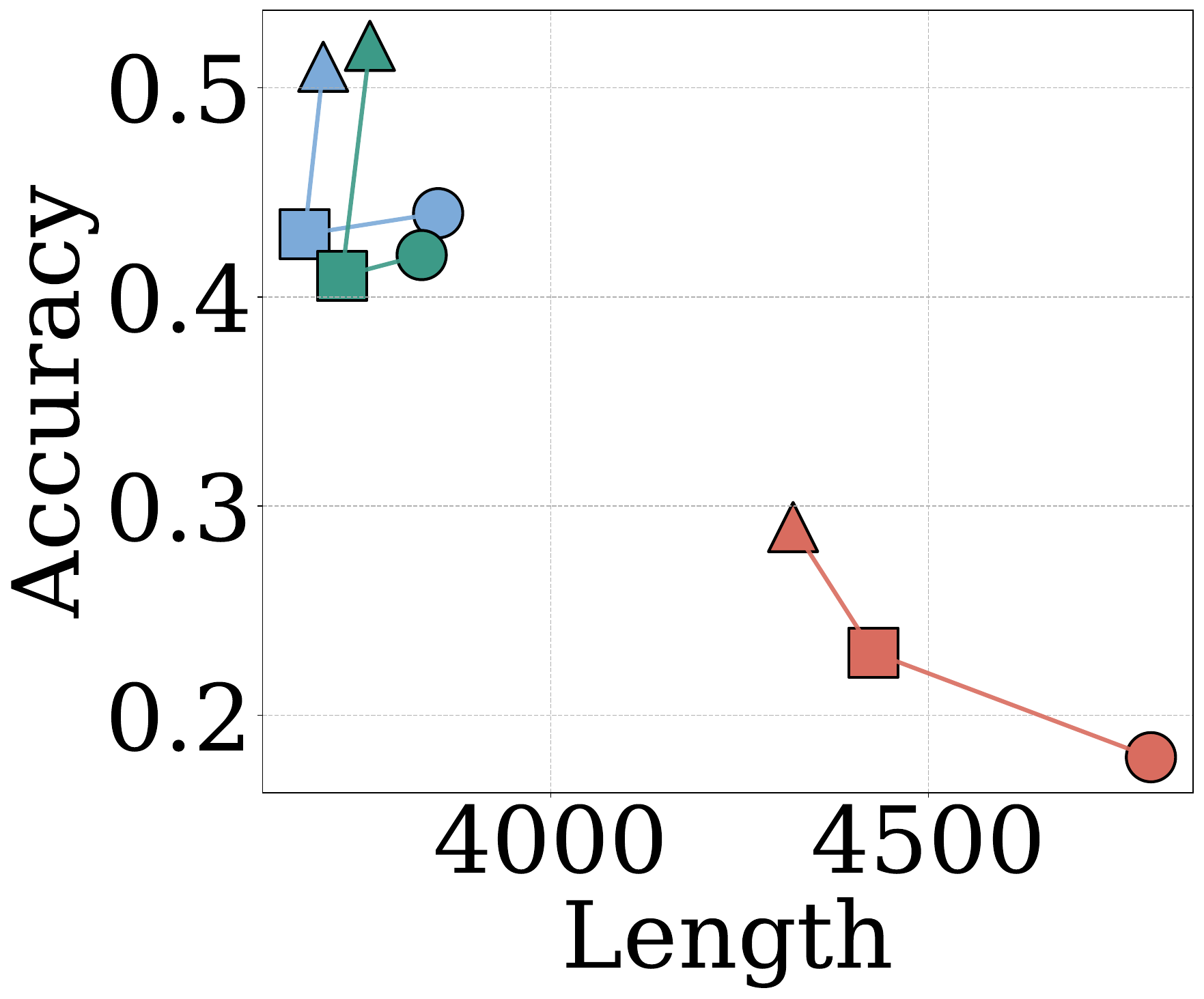} \\
        \small{(a) GPQA Diamond} & \small{(b) LiveCodeBench} \\
    \end{tabular}
    \vspace{-4mm}
    \caption{\small Accuracy vs. generation length on (a) GPQA Diamond and (b) LiveCodeBench. 
    The comparison includes the original decoding, TIP, and {\ours} on DeepSeek-R1-Distill-Qwen 1.5B/7B, and Llama 8B.
    }
    \label{fig: 2d_gpqa_code}
\end{wrapfigure}
{\ours} across models of varying sizes (1.5B, 7B, and 8B), model families (Qwen and LLaMA), and four widely used reasoning benchmarks: MATH500, AIME2024, AIME2025, and AMC2023. As we can see, {\ours} consistently improves performance over the original LRM decoding strategy across all models and datasets. For example, {DeepSeek-R1-Distill-Llama-8B} with {\ours} achieves up to a 10\% absolute accuracy gain on AIME2024, while {DeepSeek-R1-Distill-Qwen 7B} with {\ours} yields up to a 9\% improvement on AMC2023. Additionally, these accuracy gains are achieved without sacrificing the efficiency of reasoning generation: {\ours} produces comparable reasoning traces relative to the original decoding method.
\begin{wrapfigure}{r}{0.28\textwidth}
    \centering
    \vspace{-6mm}
    \includegraphics[width=0.28\textwidth]{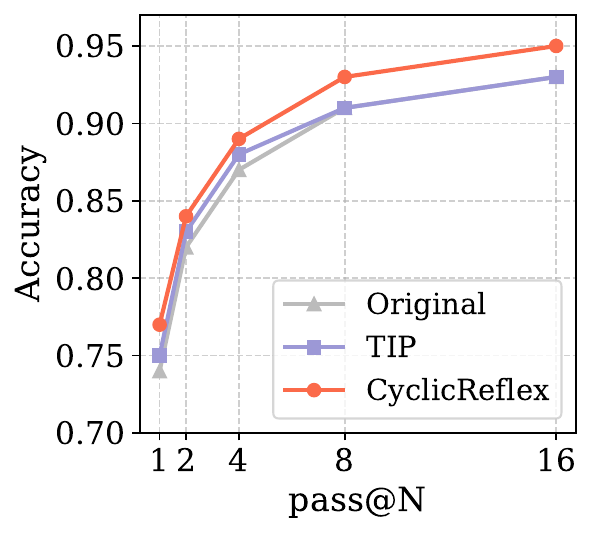}
    \vspace*{-7mm}
    \caption{\small{Pass@N of \textit{Original}, \textit{TIP}, and \textit{CyclicReflex} on MATH500 with DeepSeek-R1-Distill-Qwen-7B, evaluated at $N\!\in\!\{1,2,4,8,16\}$.}}
    \vspace*{-4mm}
    \label{fig:pass_at_n}
\end{wrapfigure}

We further compare {\ours} against two additional baselines: S1 and TIP. While S1 enforces the insertion of ``wait'' tokens at the end of each reasoning segment, leading to significantly longer outputs, it does not yield corresponding accuracy improvements. On AMC2023, in fact, S1 causes a notable performance drop, suggesting that excessive reflection may lead to overthinking. TIP, which suppresses reflection token usage, can also degrade performance in some cases. For instance, TIP causes a 3\% accuracy drop on AIME2025 when applied to {DeepSeek-R1-Distill-Llama-8B}, likely because it halts reasoning steps that are essential for solving more complex problems.

\noindent \textbf{Effectiveness of {\ours} on non-math reasoning.}
\textbf{Fig.\,\ref{fig: 2d_gpqa_code}} shows the relationship between accuracy and generation length on two \textit{non-math} benchmarks, GPQA Diamond and LiveCodeBench. Results are reported for multiple DeepSeek-R1-Distill variants (Qwen-1.5B/7B, and LLaMA-8B) under original decoding, TIP, and {\ours}. {\ours} consistently improves accuracy while maintaining response lengths comparable to TIP. In contrast, TIP can even reduce accuracy, as seen in \textbf{Fig.\,\ref{fig: 2d_gpqa_code}(b)} for DeepSeek-R1-Distill-Qwen-7B and LLaMA-8B on LiveCodeBench.


\noindent \textbf{Evaluation under Pass@N.}
In \textbf{Fig.\,\ref{fig:pass_at_n}}, we evaluate {\ours} on MATH500 with DeepSeek-R1-Distill-Qwen-7B under Pass@N. Across $N \in {1, 2, 4, 8, 16}$, {\ours} consistently achieves higher Pass@N  than Original and TIP, indicating gains that persist under multi-sample scaling. 
In addition, 
\begin{wrapfigure}{r}{0.28\textwidth}
  \centering
  \vspace*{-1mm}
  \includegraphics[width=0.28\textwidth]{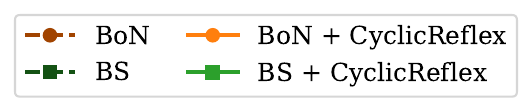}\\
  \includegraphics[width=0.28\textwidth]{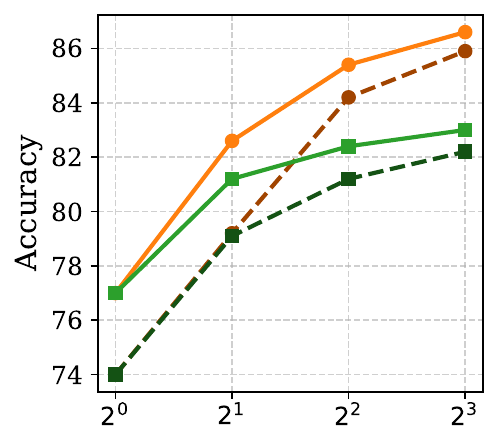}
  \vspace*{-6mm}
  \caption{\small{MATH500 accuracy of DeepSeek-R1-Distill-Qwen-1.5B under BoN/BS, w/ and w/o {\ours}.}}
  \label{fig: bon_bs}
 \vspace*{-5mm}
\end{wrapfigure}
the gain from {TIP} diminishes as $N$ increases, likely because reflection suppression reduces sampling diversity. In contrast, {\ours} maintains a stable margin as $N$ grows, better balancing cycle control with exploration.
 
\noindent \textbf{Integration with other test-time scaling methods.}
In \textbf{Fig.\,\ref{fig: bon_bs}}, we further investigate the integration of {\ours} with other test-time scaling methods across computational budgets ($2^0$ to $2^3$), using {DeepSeek-R1-Distill-Qwen-1.5B} on MATH500. We evaluate both Best-of-N (BoN) and Beam Search (BS), with generations scored using {RLHFlow-PRM-DeepSeek-8B}. Across all budget levels, BoN and Beam Search integrated with {\ours} consistently outperform their original counterparts, demonstrating the general compatibility and effectiveness of our method. Under fixed decoding strategies, BoN achieves higher accuracy than Beam Search, both with and without {\ours}. As the budget increases, the performance gap between {\ours} and the original decoding narrows, highlighting that {\ours} offers the greatest benefit under constrained inference budgets by enabling more efficient reflection token allocation.

\begin{wrapfigure}{r}{0.28\textwidth}
\centering
\vspace*{-5mm}
\includegraphics[width=0.28\textwidth]{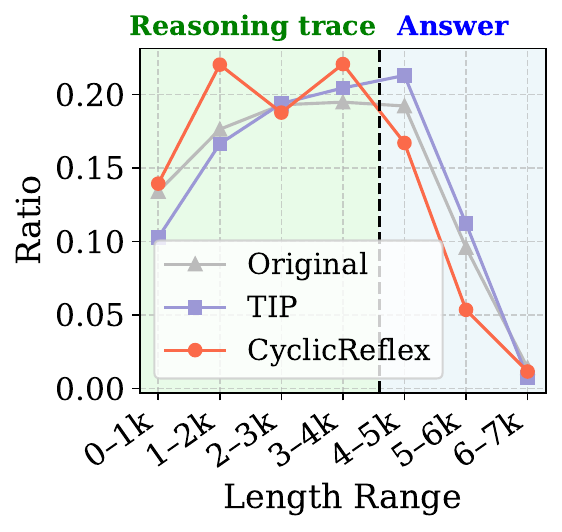}
\vspace*{-6mm}
\caption{\small{Reflection token distribution of DeepSeek-R1-Distill-Llama-8B on AIME2024 under original decoding, TIP, and {\ours}. Each curve shows the proportion of reflection tokens within 1k-token segments relative to the total generation.}}
\label{fig: line_range_ratio}
\vspace*{-3mm}
\end{wrapfigure}

\noindent \textbf{Reflection token scheduling patterns of {\ours}.}
\textbf{Fig.\,\ref{fig: line_range_ratio}} compares reflection token distributions under original decoding, TIP, and {\ours}, using DeepSeek-R1-Distill-Llama-8B on AIME2024. Each curve shows the proportion of reflection tokens within consecutive 1000-token segments relative to the total reflection count. Original decoding exhibits a gradual early rise followed by a stable, evenly spread pattern. TIP follows a similar trajectory but suppresses reflection in the 0–1k range, reflecting its tendency to inhibit early reflection. In contrast, {\ours} displays a cyclical hedging pattern with alternating peaks and troughs, allocating more reflection in the 1–2k and 3–4k ranges. This modulation avoids both excessive early suppression and late overuse, leading to stronger performance on AIME2024, improving accuracy from 0.42 (original decoding) to 0.53, demonstrating the benefit of bidirectional, position-dependent reflection scheduling.

\noindent \textbf{Improved self-correction with {\ours}.} 
We also find that {\ours} exhibits a stronger capacity for self-correction during reasoning. To evaluate this property, we select 50 incorrectly answered problems from the MATH500 dataset, originally generated by {DeepSeek-R1-Distill-Qwen-7B}. For each incorrect case, we extract the model’s reasoning trace and truncate it to three different lengths (25\%, 50\%, and 100\% of the full trace), which are then reused as misleading prompts to guide a new round of reasoning.

\begin{figure}[htb]
\vspace{-6mm}
\centering
\begin{tabular}{ccc}
\hspace*{-6mm}
\includegraphics[width=0.23\textwidth]{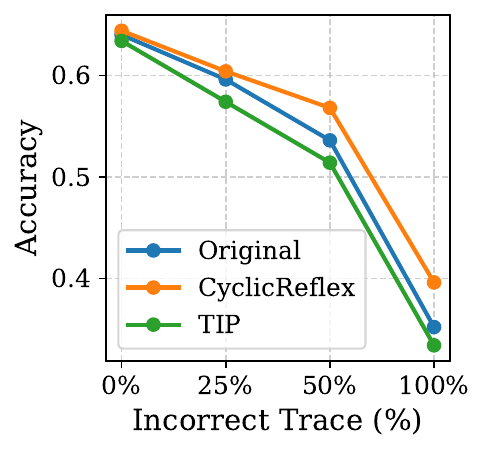} 
&
\hspace*{-5mm}
\includegraphics[width=0.24\textwidth]{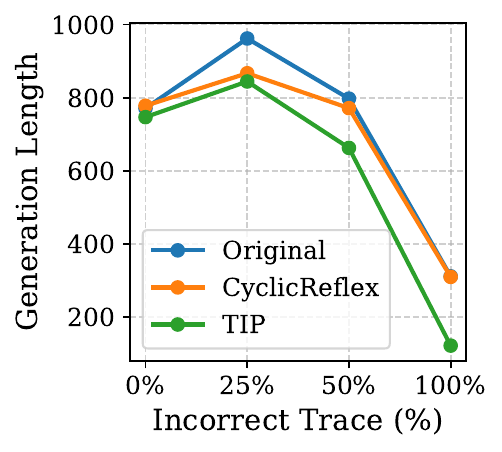}
&
\hspace*{-6mm}
\includegraphics[width=0.58\textwidth]{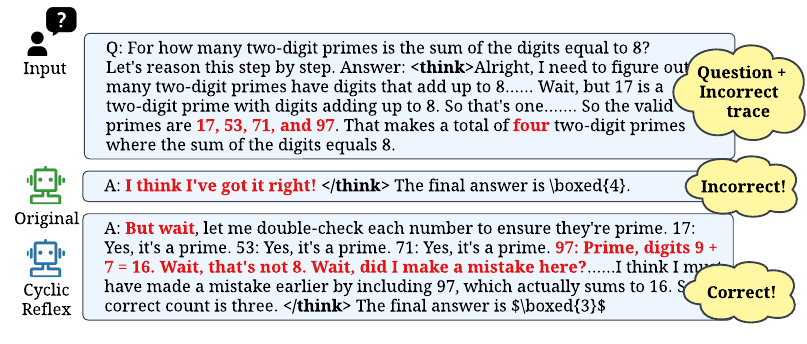}

\\
\small{\shortstack{(a) Accuracy}} 
&  
\small{\shortstack{(b) Generation length}}
&  
\small{\shortstack{(c) Example}}
\end{tabular}
\vspace{-3mm}
\caption{\small{(a)-(b) Accuracy and generation length on MATH500 with DeepSeek-R1-Distill-Qwen-7B using Original, TIP, and {\ours} decoding for questions combined with incorrect reasoning traces of different lengths. (c) Example for a question with 100\% incorrect reasoning under Original and {\ours} decoding.
}}
\vspace{-2mm}
\label{fig: self-correction}
\end{figure}

Under each prompt condition, we prompt the same model ({DeepSeek-R1-Distill-Qwen-7B}) to re-answer the question five times and report the average accuracy. As shown in \textbf{Fig.\,\ref{fig: self-correction}(a)}, {\ours} significantly outperforms both the original decoding and the TIP baseline across all trace lengths. 

Notably, the performance gap increases with longer misleading traces, indicating that {\ours} enhances the model’s ability to resist and correct earlier reasoning errors. 
In addition, this improved self-correction does not come at the cost of generation efficiency. As shown in \textbf{Fig.\,\ref{fig: self-correction}(b)}, the generation lengths under {\ours} remain comparable to those of the original decoding strategy.
In contrast, TIP tends to suppress reflection token usage, which hampers the model’s ability to re-evaluate its own reasoning and results in markedly lower accuracy.
Finally, \textbf{Fig.\,\ref{fig: self-correction}(c)} provides a concrete example. Given a MATH500 question with a 100\% incorrect reasoning trace in which the model incorrectly claims that the digits of 17, 53, 71, and 97 sum to 8, the original decoding strategy fails to correct the error and outputs the wrong answer, 4. In contrast, {\ours} initiates a double-check, correctly identifies the error (specifically excluding 97), and ultimately outputs the correct answer, 3.

\begin{wrapfigure}{r}{.45\textwidth}
    \vspace*{-4mm}
    \centering
    \begin{tabular}{cc} 
        \hspace*{-2mm} 
        \includegraphics[width=0.22\textwidth] {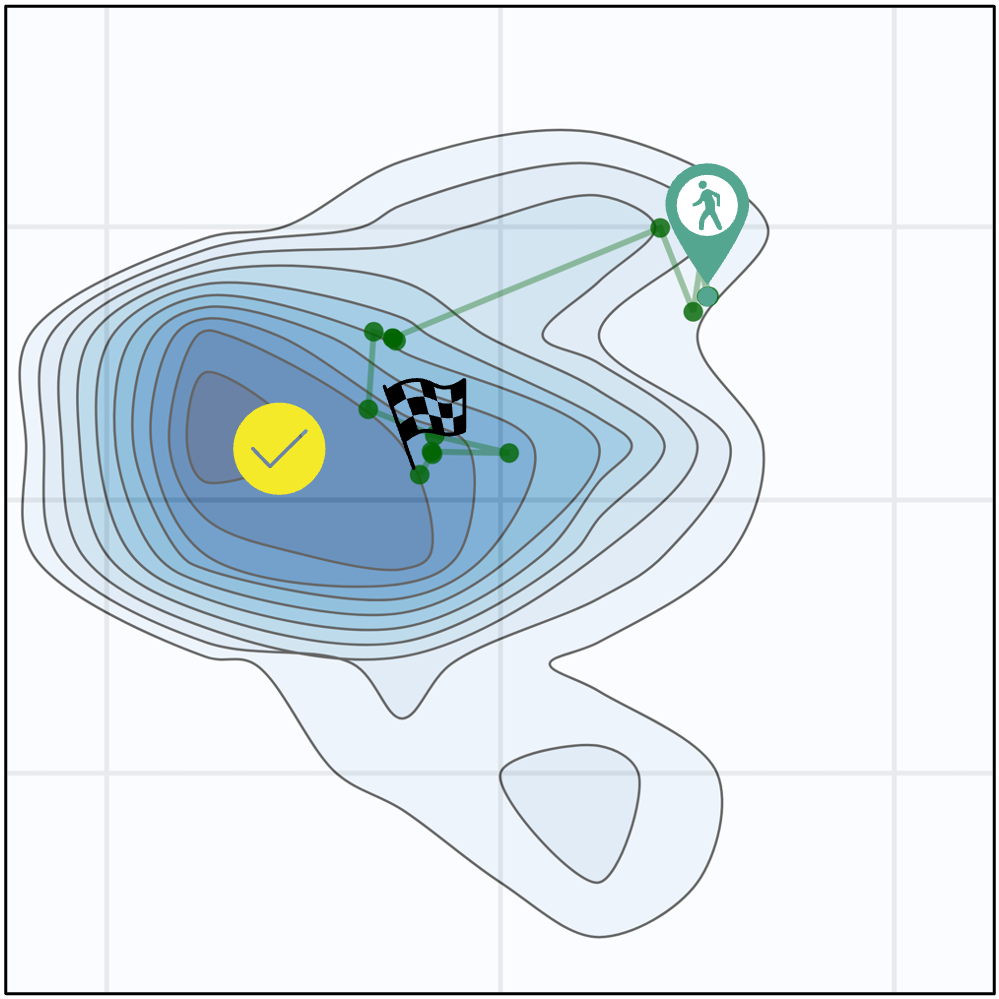} &   \hspace*{-2mm}
        \includegraphics[width=0.22\textwidth]
        {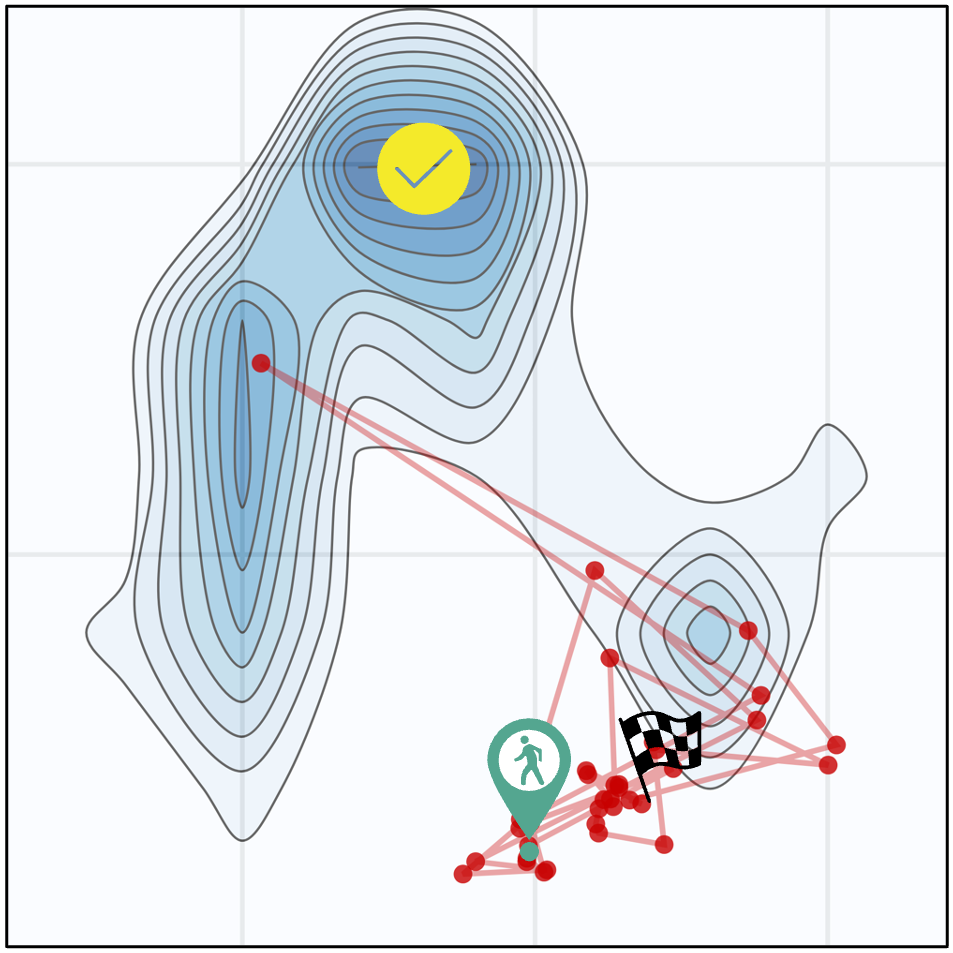} \\
        \hspace*{-2mm} \small{(a) {\ours}} & \hspace*{-2mm}
        \small{(b) Original}
    \end{tabular}
    \caption{\small{Examples of landscape of thought for DeepSeek-R1-Distill-Qwen-7B with original and {\ours} decoding strategy. The format follows Fig\,\ref{fig: lot}.
    }}
    \label{fig: cr_lot}
    \vspace*{-4mm}
\end{wrapfigure}

\noindent \textbf{Visualizing {\ours}'s efficacy using the landscape of thoughts.} In \textbf{Fig.\,\ref{fig: cr_lot}}, we present the landscape of thoughts generated by {DeepSeek-R1-Distill-Qwen-7B} under the original decoding and {\ours}. The darkest area in the landscape represents the correct answer, while other dark regions correspond to misleading areas that may attract the reasoning trajectory toward alternative but incorrect answers. As shown in \textbf{Fig.\,\ref{fig: cr_lot}(a)}, the landscape produced by {\ours} is more concentrated, with fewer distracting regions. This suggests that the reasoning trajectory remains focused and is more likely to converge directly to the correct answer.

In contrast, the landscape under the original decoding strategy in \textbf{Fig.\,\ref{fig: cr_lot}(b)} is more scattered, making the model more susceptible to misleading areas and eventually reaching an incorrect final answer. This comparison demonstrates that our method enhances reasoning efficiency by reducing unnecessary detours and guiding the model toward the correct solution path more effectively.

\noindent \textbf{Additional analyses and ablations.}
As shown in \textbf{Fig.\,\ref{fig: tip_cr_improvement}} of Appendix\,\ref{appendix: add_exp}, {\ours} improves MATH500 accuracy across all difficulty levels, whereas TIP mainly benefits the Hard split. \textbf{Table\,\ref{tab: dynamic_reflection_set}} shows that {\ours} is robust to the choice of reflection tokens; and
\textbf{Table\,\ref{tab: adjustment_baselines}} confirms that {\ours} outperforms alternative reflection-logit adjustments, supporting the necessity of our cyclic hedging schedule. Finally, \textbf{Table\,\ref{tab: cr_example}} in \textbf{Appendix\,\ref{appendix: example}} provides  generation examples comparing the original decoding with {\ours}.

%% file: tab/model_comparison.tex
\begin{table}[htbp]
\vspace{-6mm}
\caption{\small{Accuracy (Acc) and generation length (Len) comparison on four math reasoning benchmarks (MATH500, AIME2024, AIME2025, and AMC2023) using {DeepSeek-R1-Distilled Model: Qwen 1.5B, Qwen 7B, and Llama 8B}. Each model is evaluated under four decoding strategies: \textit{Original}, \textit{TIP}, \textit{S1}, and \textit{\ours}. The best accuracy in each setting is highlighted in \textbf{bold}, while the second-best is \underline{underlined}.}}
\label{tab: model_comparison}
\vspace{2mm}
\begin{center}
\renewcommand{\arraystretch}{1.0}
\resizebox{0.7\textwidth}{!}{
\begin{tabular}{c|cc|cc|cc|cc}
\toprule[1pt]
\midrule
\multirow{2}{*}{\textbf{Method}}
& \multicolumn{2}{c|}{\textbf{MATH500}} 
& \multicolumn{2}{c|}{\textbf{AIME2024}} 
& \multicolumn{2}{c|}{\textbf{AIME2025}} 
& \multicolumn{2}{c}{\textbf{AMC2023}} \\
\cline{2-9}
& Acc & Len & Acc & Len & Acc & Len & Acc & Len \\
\midrule
\multicolumn{9}{c}{\textbf{DeepSeek-R1-Distill-Qwen-1.5B}} \\
\midrule
Original & 0.74 & 1253.05 & \underline{0.23} & 3584.36 & 0.19 & 3442.07 & \underline{0.63} & 1855.85 \\
TIP      & \underline{0.75} & 1206.91 & \underline{0.23} & 3329.17 & \underline{0.20} & 3825.17 & \underline{0.63} & 1890.35 \\
S1       & 0.73 & 1532.05 & 0.17 & 4112.07 & \underline{0.20} & 3867.71 & 0.45 & 3263.75 \\
\rowcolor{Gray}
\ours    & \textbf{0.77} & 1212.94 & \textbf{0.30} & 3547.10 & \textbf{0.23} & 3467.97 & \textbf{0.65} & 1839.23 \\
\midrule
\multicolumn{9}{c}{\textbf{DeepSeek-R1-Distill-Qwen-7B}} \\
\midrule
Original & 0.86 & 785.25 & \underline{0.43} & 2878.39 & 0.31 & 3192.59 & 0.81 & 1300.53 \\
TIP      & \underline{0.87} & 775.77 & \underline{0.43} & 2806.53 & 0.30 & 3107.30 & \underline{0.85} & 1267.83 \\
S1       & 0.83 & 1190.96 & 0.33 & 3541.10 & \underline{0.33} & 3455.33 & \underline{0.85} & 2158.00 \\
\rowcolor{Gray}
\ours    & \textbf{0.89} & 777.93 & \textbf{0.50} & 2868.30 & \textbf{0.37} & 3190.33 & \textbf{0.90} & 1229.25 \\
\midrule
\multicolumn{9}{c}{\textbf{DeepSeek-R1-Distill-Llama-8B}} \\
\midrule
Original & \underline{0.83} & 1196.98 & 0.42 & 3593.73 & \underline{0.30} & 3922.41 & 0.81 & 1951.88 \\
TIP      & \underline{0.83} & 1080.62 & \underline{0.47} & 3572.40 & 0.27 & 3866.00 & \underline{0.85} & 1932.63 \\
S1       & 0.78 & 1461.93 & 0.43 & 3742.27 & 0.27 & 4351.87 & 0.75 & 2812.75 \\
\rowcolor{Gray}
\ours    & \textbf{0.85} & 1108.30 & \textbf{0.53} & 3454.97 & \textbf{0.37} & 3856.80 & \textbf{0.90} & 1942.40 \\
\midrule
\bottomrule[1pt]
\end{tabular}}
\renewcommand{\arraystretch}{1.0}
\vspace{-2mm} 
\end{center}
\end{table}

%% file: sections/conclusion.tex
\section{Conclusion}
\label{sec: conclusion}
We introduce the problem of resource allocation in LRMs, focusing on the challenge of managing reflection tokens during test-time generation. We show that both under-reflection and over-reflection, stemming from insufficient or excessive use of reflection tokens, can severely degrade reasoning performance.
To address this, we draw a conceptual analogy between reflection token scheduling and learning rate control in optimization, and propose {\ours}, a training-free decoding strategy that cyclically modulates reflection token logits using a triangular waveform. 
{\ours} dynamically adapts to the evolving stage of reasoning, enabling more balanced token allocation.
Extensive experiments across multiple reasoning benchmarks demonstrate that {\ours} consistently improves accuracy, enhances self-correction capability, and integrates seamlessly with existing test-time scaling methods. Our work highlights the critical role of reflection tokens as a valuable resource for LRMs and opens new avenues for principled, adaptive reasoning control. The use of LLM, limitations and broader impacts are further discussed in \textbf{Appendix\,\ref{appendix: llm_usage}}, \textbf{Appendix\,\ref{appendix: limitation}} and \textbf{Appendix\,\ref{appendix: impact}}.


%% file: sections/acknowledgment.tex
\clearpage\newpage
\section*{Acknowledgment}
This research is supported in part by the ARO Award W911NF2310343.  
Additionally, the contributions of C.~Fan, Y.~Zhang, J.~Jia, and S.~Liu are partially supported by the NSF CISE Core Program Awards IIS-2207052 and IIS-2504263, the Open Philanthropy Research Award, and the Schmidt Sciences' Trustworthy AI Program. The contributions of A. Hero were also supported by the
United States National Science Foundation under Grant NSF CCF 2246213.

%% file: sections/appendix.tex
\clearpage\newpage
\onecolumn
\section*{\Large{Appendix}}
\setcounter{section}{0}
\setcounter{figure}{0}
\setcounter{table}{0}
\makeatletter 
\renewcommand{\thesection}{\Alph{section}}
\renewcommand{\theHsection}{\Alph{section}}
\renewcommand{\thefigure}{A\arabic{figure}} 
\renewcommand{\theHfigure}{A\arabic{figure}} 
\renewcommand{\thetable}{A\arabic{table}}
\renewcommand{\theHtable}{A\arabic{table}}
\makeatother

\renewcommand{\thetable}{A\arabic{table}}
\setcounter{mylemma}{0}
\renewcommand{\themylemma}{A\arabic{mylemma}}
\setcounter{equation}{0}
\renewcommand{\theequation}{A\arabic{equation}}

\input{sections/appendix/exp_setup}
\input{sections/appendix/add_exp}
\input{sections/appendix/example}
\clearpage\newpage
\input{sections/appendix/usage}
\input{sections/appendix/limitation}
\input{sections/appendix/impact}

%% file: sections/appendix/exp_setup.tex
\section{Detailed Experiment Setups}
\label{appendix: exp_setup}
\subsection{Computing Resources}
All experiments are conducted on a single node equipped with 8 NVIDIA A6000 GPUs.

\subsection{Decoding Details}

During generation, we employ the vLLM framework to enable efficient and scalable inference. The decoding configuration is as follows: the maximum number of new tokens is set to 8192; the top-$p$ value is set to 0.95; and the temperature is set to 0.6. Top-$p$ sampling selects tokens from the smallest possible set whose cumulative probability exceeds $p$, effectively filtering out low-probability candidates to maintain generation quality while allowing diversity. A temperature of 0.6 sharpens the token probability distribution, promoting more deterministic and focused outputs by reducing sampling randomness.

For {\ours} on the MATH500 and AMC2023 datasets, we perform a grid search over $A \in [1, 10]$ and $C \in [200, 1000]$. On the AIME2024 and AIME2025 datasets, we perform a grid search over $A \in [1, 10]$ and $C \in [1000, 2000]$. For TIP, we conduct a grid search with $\alpha \in [-10, -1]$ and $T_0 \in [100, 1000]$. For S1, we forcefully insert the reflection token ``Wait'' after the model generates </think>, prompting continued reasoning.

In the Best-of-N setting, the LRM generates multiple independent candidate answers, and the PRM selects the most preferred one based on final-answer evaluation. For Beam Search, we perform a step-by-step search guided by PRM feedback to optimize cumulative reward. Throughout decoding, we use multiple candidate beams with a fixed beam width of 4.

\subsection{Prompt Details}
We present the prompt used to evaluate the reasoning ability of the LRM. For each question, we replace the \texttt{\{question\}} placeholder in the User section of the prompt. After the Assistant generates the reasoning trace and the final answer (\texttt{\{generation\}}), we follow the approach of~\citet{yang2024qwen2} to first extract the final answer and then apply rule-based matching to assess its correctness.

\begin{prompt}{Evaluation prompt}\label{prompt:meta}
\textbf{System}:\\
You are a helpful AI bot that answers questions for a user. Keep your response short and direct.\\ \\
\textbf{User}:\\
Question: \{question\} \\ Let's reason this step by step.\\ \\
\textbf{Assistant}:\\
Answer: \{generation\}
\end{prompt}

%% file: sections/appendix/add_exp.tex
\section{Additional Experiment Results}
\label{appendix: add_exp}

\paragraph{Ablation study on {\ours}'s hyperparameter.} In \textbf{Fig.\,\ref{fig: hyper}}, we study how {\ours}’s key hyperparameters affect final performance. Following Fig.\,\ref{fig: cyclic-plot}, we vary three parameters: the amplitude $A$, the period $C$, and the initial phase shift $\phi$.
\textbf{Fig.\,\ref{fig: hyper}(a)} analyzes $A$ and $C$ by reporting the number of additional correct answers over the original decoding strategy (\textit{improvement}). Overall, $C$ has a larger effect than $A$; in particular, $C=600$ achieves the best accuracy across all tested amplitudes ($A\in\{4.0,5.0,6.0\}$).
Finally, \textbf{Fig.\,\ref{fig: hyper}(b)} shows that $\phi=0$ performs best (matching the schedule in Fig.\,\ref{fig: cyclic-plot}), suggesting that promoting reflection tokens early can help, while gradually suppressing their logits later improves convergence.

\begin{figure}[htb]
\centering
\begin{tabular}{cc}
\hspace*{-6mm}
\includegraphics[width=0.22\textwidth]{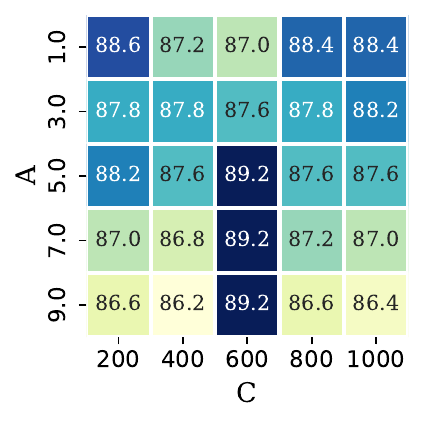} 
&
\hspace*{-4mm}
\includegraphics[width=0.24\textwidth]{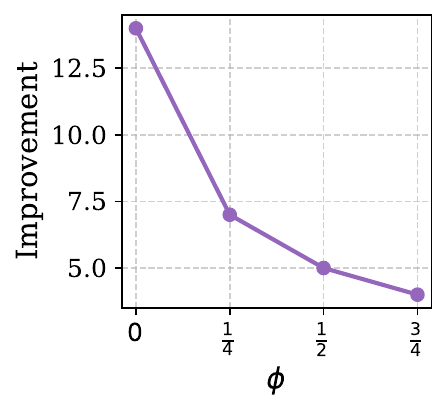}
\\
\small{\shortstack{(a) Amplitude and period}} 
&  
\hspace*{4mm}
\small{\shortstack{(b) Shift}}
\end{tabular}
\caption{\small{
(a) Improvement heatmaps of DeepSeek-R1-Distill-Qwen-7B on MATH500 across amplitude $A$ and period $C$, measured as the number of additional correct answers. (b) Improvement of {\ours} over the original decoding strategy under different initial phase shifts $\phi$.
}}
\label{fig: hyper}
\end{figure}

\begin{figure}[htbp]
  \centering
  \includegraphics[width=0.5\textwidth]{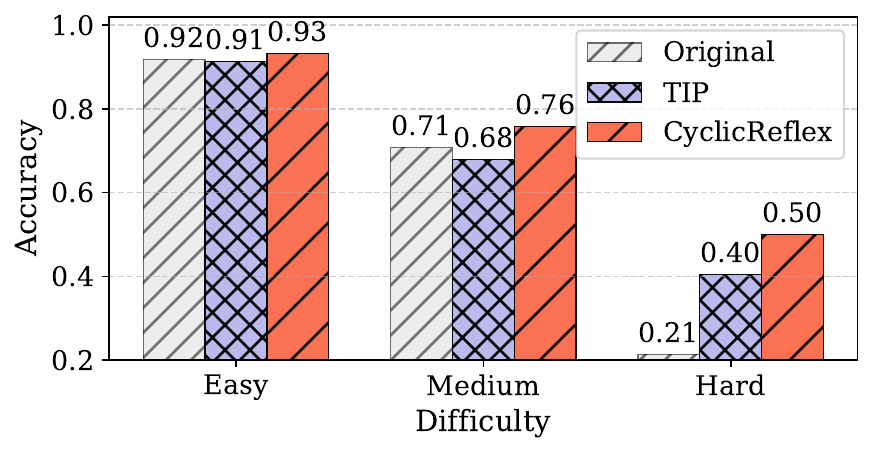}
  \vspace{-2mm}
  \caption{\small{
    Improvement of DeepSeek-R1-Distill-Qwen-7B on MATH500 by TIP and {\ours} across difficulty levels (Easy, Medium, Hard), following Fig.\,\ref{fig: tip_motivation} (a) and (b).
  }}
  \label{fig: tip_cr_improvement}
\end{figure}

\paragraph{Accuracy of {\ours} at different difficulty levels on MATH500.}
In \textbf{Fig.\,\ref{fig: tip_cr_improvement}}, we categorize the MATH500 dataset by difficulty level to closely examine where the accuracy improvements from {\ours} are most pronounced. The grouping strategy follows that used in Fig.\,\ref{fig: tip_motivation}(a), and the accuracy is reported in a manner consistent with Fig.\,\ref{fig: tip_motivation}(b).
For comparison, we also include TIP as a baseline. We observe that {\ours} consistently improves accuracy across all difficulty levels (Easy, Medium, and Hard) whereas TIP primarily yields gains on Hard problems and even leads to performance degradation on Easy and Medium ones. This contrast stems from the bidirectional nature of {\ours}, which allows for dynamic promotion or suppression of reflection token sampling based on the current stage of the reasoning process. Such flexibility enables {\ours} to better adapt to problem difficulty, leading to more effective resource allocation and improved overall performance.

\paragraph{Generalization to other model families and scales.}
We further extend {\ours} to a different model family  by evaluating \texttt{Qwen3-4B}, \texttt{Qwen3-8B} and \texttt{Qwen3-14B} on {MATH500}, {AIME2024}, {AIME2025}, {AMC2023} and {LiveCodeBench}, with \textit{Original} and \textit{TIP} included as baselines. The results are provided in \textbf{Table\,\ref{tab: qwen}}. Across all five benchmarks, {\ours} consistently outperforms both baselines, even as the underlying model scale increases and the family differs from the DeepSeek-distilled lineage. This demonstrates that our method generalizes beyond a single family of reasoning models and remains robust and effective for a {larger} model with an {alternative} architecture.

\begin{table}[htbp]
\vspace{-2mm}
\caption{\small{Accuracy comparison on five benchmarks (MATH500, AIME2024, AIME2025, AMC2023, and LiveCodeBench) using Qwen3-4B, Qwen3-8B, and Qwen3-14B under three decoding strategies: \textit{Original}, \textit{TIP}, and \textit{CyclicReflex}.}}
\label{tab: qwen}
\begin{center}
\renewcommand{\arraystretch}{1.0}
\resizebox{0.65\textwidth}{!}{
\begin{tabular}{c|ccccc}
\toprule[1pt]
\midrule
{\textbf{Method}}
& \multicolumn{1}{c}{\textbf{MATH500}}
& \multicolumn{1}{c}{\textbf{AIME2024}}
& \multicolumn{1}{c}{\textbf{AIME2025}}
& \multicolumn{1}{c}{\textbf{AMC2023}}
& \multicolumn{1}{c}{\textbf{LiveCodeBench}} \\
\midrule
\multicolumn{6}{c}{\textbf{Qwen3-4B}} \\
\midrule
Original     & 0.60 & 0.13 & 0.03 & 0.47 & 0.58 \\
TIP          & 0.63 & 0.17 & 0.07 & 0.49 & 0.59 \\
\rowcolor{Gray}
CyclicReflex & 0.68 & 0.23 & 0.10 & 0.53 & 0.61 \\
\midrule
\multicolumn{6}{c}{\textbf{Qwen3-8B}} \\
\midrule
Original     & 0.83 & 0.37 & 0.27 & 0.79 & 0.59 \\
TIP          & 0.84 & 0.40 & 0.23 & 0.80 & 0.60 \\
\rowcolor{Gray}
CyclicReflex & 0.86 & 0.43 & 0.33 & 0.84 & 0.63 \\
\midrule
\multicolumn{6}{c}{\textbf{Qwen3-14B}} \\
\midrule
Original     & 0.88 & 0.43 & 0.33 & 0.83 & 0.63 \\
TIP          & 0.88 & 0.47 & 0.33 & 0.82 & 0.63 \\
\rowcolor{Gray}
CyclicReflex & 0.90 & 0.50 & 0.40 & 0.88 & 0.67 \\
\midrule
\bottomrule[1pt]
\end{tabular}}
\renewcommand{\arraystretch}{1.0}
\vspace{-4mm}
\end{center}
\end{table}

\paragraph{Robustness to the reflection-token set.}
Next, we show that the proposed {\ours} is robust to the choice of reflection tokens and can even operate with a dynamically updated reflection-token set. To validate this, we consider a dynamic variant in which the reflection-token set is expanded online during decoding: we start from a small seed set (\textit{e.g.}, ``wait'', ``but''), and at decoding step $t$, if the Top-1 token belongs to the current reflection-token set and the logit gap between Top-2 and Top-1 is smaller than the gap between Top-2 and Top-3, we add the Top-2 token to the set. The rationale is that the model tends to use in contexts similar to existing reflection tokens, gradually enriching the set and yielding additional reflection-like tokens (\textit{e.g.}, ``hmm'', ``maybe'', ``perhaps''), which are absent from the original reflection-token set. We evaluate {\ours} with both fixed and dynamic reflection-token sets on DeepSeek-R1-Distill-Qwen-7B across MATH500, AIME2024, AIME2025, and AMC2023, and compare against the Original decoding and TIP. As shown in \textbf{Table\,\ref{tab: dynamic_reflection_set}}, the two variants perform comparably across all benchmarks, indicating that {\ours} exhibits robustness to the selection of reflection tokens.

\begin{table}[htbp]
\centering
\caption{\small{Accuracy of \textit{Original}, \textit{TIP}, and \textit{\ours} variants on MATH500, AIME2024, AIME2025, and AMC2023 with DeepSeek-R1-Distill-Qwen-7B.}}
\label{tab: dynamic_reflection_set}
\vspace{1mm}
\renewcommand{\arraystretch}{1.0}
\resizebox{0.65\textwidth}{!}{
\begin{tabular}{c|c|c|c|c}
\toprule[1pt]
\toprule
\textbf{Method} & \textbf{MATH500} & \textbf{AIME2024} & \textbf{AIME2025} & \textbf{AMC2023} \\
\midrule
Original      & 0.86              & \underline{0.43}   & \underline{0.31}   & 0.81              \\
TIP           & \underline{0.87}  & \underline{0.43}   & 0.30               & 0.85              \\
CyclicReflex w/ dynamic set
              & \textbf{0.89}     & \textbf{0.50}      & \textbf{0.37}      & \textbf{0.91}     \\
\rowcolor{Gray}
Ours: CyclicReflex
              & \textbf{0.89}     & \textbf{0.50}      & \textbf{0.37}      & \underline{0.90}   \\
\bottomrule
\bottomrule[1pt]
\end{tabular}}
\renewcommand{\arraystretch}{1.0}
\end{table}

\paragraph{Comparison with approaches using additional reflection-logit adjustments.}
In \textbf{Table\,\ref{tab: adjustment_baselines}}, we extend our comparison on {DeepSeek-R1-Distill-Qwen-7B} by adding several additional adjustment-based baselines: (1) \textit{TIP w/ positive adjustment}, which always boosts reflection-token logits rather than penalizing them; (2) \textit{random adjustment}, which injects zero-mean noise into the reflection-token logits at every decoding step; and (3) \textit{linear decay}, which begins with a strong positive boost that linearly decreases into a penalty across the generation process. 
Across {MATH500}, {AIME2024}, {AIME2025}, and {AMC2023}, {\ours}  consistently achieves the highest accuracy. In contrast, both the positive-adjustment and random-adjustment variants underperform even standard TIP, and the linear-decay baseline narrows, but does not close, the performance gap relative to our method.
These findings indicate that naïve or purely monotonic adjustments to reflection-token logits are insufficient, and that the \textit{cyclic hedging schedule} in {\ours} provides a fundamentally more effective mechanism for balancing exploration and control during reasoning.

\begin{table}[htbp]
\centering
\caption{\small{Performance of \textit{Original}, \textit{TIP}, and adjustment-based variants on MATH500, AIME2024, AIME2025, and AMC2023 with DeepSeek-R1-Distill-Qwen-7B.}}
\label{tab: adjustment_baselines}
\vspace{1mm}
\renewcommand{\arraystretch}{1.0}
\resizebox{0.65\textwidth}{!}{
\begin{tabular}{c|c|c|c|c}
\toprule[1pt]
\toprule
\textbf{Method} & \textbf{MATH500} & \textbf{AIME2024} & \textbf{AIME2025} & \textbf{AMC2023} \\
\midrule
Original                  & 0.86              & 0.43              & 0.31              & 0.81              \\
TIP                       & \underline{0.87}  & 0.43              & 0.30              & 0.85              \\
TIP w/ positive adjustment& 0.85              & 0.40              & \underline{0.33}  & 0.83              \\
Random adjustment         & 0.82              & 0.41              & 0.30              & 0.79              \\
Linear decay              & \underline{0.87}  & \underline{0.45}  & \underline{0.33}  & \underline{0.87}  \\
\rowcolor{Gray}
Ours: CyclicReflex        & \textbf{0.89}     & \textbf{0.50}     & \textbf{0.37}     & \textbf{0.90}     \\
\bottomrule
\bottomrule[1pt]
\end{tabular}}
\renewcommand{\arraystretch}{1.0}
\end{table}

%% file: sections/appendix/example.tex
\section{Generation Examples}
\label{appendix: example}


In \textbf{Table\,\ref{tab: cr_example}-\ref{tab: cr_example_4}}, we present the answers generated for questions in MATH500 under both the original decoding strategy and {\ours}. For better comparison, we also include the ground truth labels. Compared to the original decoding strategy, {\ours} demonstrates deeper reasoning and avoids prematurely terminating the thinking process. For instance, in Q1, the original answer is given as $\binom{14}{5} - \binom{8}{5} - \binom{6}{5}$, whereas {\ours} continues simplifying the expression to arrive at the final result. Moreover, {\ours} exhibits a better understanding of the problem semantics. In Q2, for example, the original decoding misinterprets the notion of projection, leading to an incorrect answer. Finally, {\ours} is more adept at identifying key problem elements. In Q3, the original response overlooks the trap involving $x = 9$, while {\ours} correctly accounts for this subtlety.

\begin{table*}[htb]
\begin{center}
\caption{\small{Examples from DeepSeek-R1-Distill-Qwen-7B on MATH500 under the original decoding strategy and CyclicReflex. Reflection tokens are highlighted in \textcolor{red}{red}.}}
\resizebox{1.0\textwidth}{!}{
\begin{tabular}{c|p{18cm}}
\toprule[1pt] \midrule
\multicolumn{2}{c}{\makecell[c]{\textbf{Q1:} A Senate committee has 8 Republicans and 6 Democrats. In how many ways \\ can we form a subcommittee of 5 members that has at least one member from each party?}} \\
\midrule
\textbf{Label:} & 1940 \\
\textbf{Original:} &
<think> Okay, so I have this problem where there's a Senate committee with 8 Republicans and 6 Democrats. I need to figure out how many ways we can form a subcommittee of 5 members that has at least one member from each party. Hmm, let's break this down.

First, I know that the total number of people on the committee is 8 + 6, which is 14. If I were to form a subcommittee of 5 without any restrictions, the number of ways would just be the combination of 14 taken 5 at a time, right? So that's C(14,5). \textcolor{red}{But} the problem says there has to be at least one from each party, so I can't have all Republicans or all Democrats in the subcommittee.

So, I think I need to subtract the cases where the subcommittee is all Republicans or all Democrats. Let me calculate those separately. For all Republicans, it's C(8,5) because there are 8 Republicans. For all Democrats, it's C(6,5) since there are 6 Democrats.

Putting it all together, the total number of valid subcommittees should be C(14,5) minus C(8,5) minus C(6,5). Let me make sure that makes sense. By subtracting the all-Republican and all-Democrat cases from the total, I'm left with only those subcommittees that have at least one from each party. Yeah, that seems right.

I should probably calculate these values to get the exact number, \textcolor{red}{but} I think the reasoning is solid. So the formula is C(14,5) - C(8,5) - C(6,5). I'll go with that.

</think>

The number of ways is calculated by subtracting the all-Republican and all-Democrat subcommittees from the total.

Total ways: \( \binom{14}{5} - \binom{8}{5} - \binom{6}{5} \)

\textbf{Answer}: \( \boxed{\dbinom{14}{5} - \dbinom{8}{5} - \dbinom{6}{5}} \) \\
\textbf{{\ours}:} & 
<think> Okay, so I have this problem where there's a Senate committee with 8 Republicans and 6 Democrats. I need to figure out how many ways we can form a subcommittee of 5 members that has at least one member from each party. Hmm, let's break this down.

First, I know that the total number of people on the committee is 8 + 6, which is 14. If I were to form a subcommittee of 5 without any restrictions, the number of ways would just be the combination of 14 taken 5 at a time, right? So that's C(14,5). \textcolor{red}{But} the problem says there has to be at least one from each party, so I can't have all Republicans or all Democrats in the subcommittee.

So, I think I need to subtract the cases where the subcommittee is all Republicans or all Democrats. Let me calculate those separately. For all Republicans, it's C(8,5) because there are 8 Republicans. For all Democrats, it's C(6,5) since there are 6 Democrats.

\textcolor{red}{But} hold on, I should make sure these numbers are correct. C(8,5) is the number of ways to choose 5 Republicans out of 8, and C(6,5) is choosing 5 Democrats out of 6. I think that's right.

So, the total number of valid subcommittees would be C(14,5) minus C(8,5) minus C(6,5). Let me compute those values.

C(14,5) is 2002, C(8,5) is 56, and C(6,5) is 6. So, 2002 - 56 - 6 equals 1940. That should be the number of ways to form the subcommittee with at least one from each party.

</think>

The number of ways to form a subcommittee of 5 members with at least one member from each party is calculated by subtracting the all-Republican and all-Democrat subcommittees from the total number of possible subcommittees.

- Total subcommittees: \( \binom{14}{5} = 2002 \)

- All-Republican subcommittees: \( \binom{8}{5} = 56 \)

- All-Democrat subcommittees: \( \binom{6}{5} = 6 \)

Thus, the number of valid subcommittees is \( 2002 - 56 - 6 = 1940 \).

\textbf{Answer}: 1940 \\
\midrule
\bottomrule[1pt]
\end{tabular}
}
\label{tab: cr_example}
\end{center}
\end{table*}

\begin{table*}[htb]
\begin{center}
\caption{\small{Examples from DeepSeek-R1-Distill-Qwen-7B on MATH500 under the original decoding strategy and {\ours}. Reflection tokens are highlighted in \textcolor{red}{red}.}}
\resizebox{1.0\textwidth}{!}{
\begin{tabular}{c|p{18cm}}
\toprule[1pt] \midrule
\multicolumn{2}{c}{\makecell[c]{\textbf{Q2:} Find the projection of \( \mathbf{a} \) onto \( \mathbf{b} = \begin{pmatrix} 2 \\\\ 6 \\\\ 3 \end{pmatrix} \), if \( \mathbf{a} \cdot \mathbf{b} = 8 \).}} \\
\midrule
\textbf{Label:} & \( \begin{pmatrix} \dfrac{16}{49} \\\\ \dfrac{48}{49} \\\\ \dfrac{24}{49} \end{pmatrix} \) \\
\textbf{Original:} & 
<think> Okay, so I need to find the projection of vector $\mathbf{a}$ onto vector $\mathbf{b}$. I remember that the formula for the projection of $\mathbf{a}$ onto $\mathbf{b}$ is $(\mathbf{a} \cdot \mathbf{b})$ divided by the magnitude of $\mathbf{b}$ squared, and then multiplied by vector $\mathbf{b}$. \textcolor{red}{But} \textcolor{red}{wait}, in this case, they only gave me that the dot product of $\mathbf{a}$ and $\mathbf{b}$ is $8$. Hmm, so maybe I don't need the entire vector $\mathbf{b}$, just the scalar projection?

Let me think. The formula for the scalar projection is $(\mathbf{a} \cdot \mathbf{b}$ divided by the magnitude of $\mathbf{b}$. Since they already gave me $\mathbf{a} \cdot \mathbf{b} = 8$, I just need to find the magnitude of $\mathbf{b}$. The vector $\mathbf{b}$ is $[2, 6, 3]$, so its magnitude is the square root of $(2^2 + 6^2 + 3^2)$. Calculating that, $2^2 = 4$, $6^2 = 36$, and $3^2 = 9$. Adding those up gives $4 + 36 + 9 = 49$. The square root of $49$ is $7$. So the magnitude of $\mathbf{b}$ is $7$.

Therefore, the scalar projection of $\mathbf{a}$ onto $\mathbf{b}$ is $8$ divided by $7$, which is $8/7$. I think that's it. The problem didn't specify whether they wanted the vector projection or just the scalar, \textcolor{red}{but} since they only gave me a scalar product, I guess it's the scalar projection they're asking for. So the answer should be $8/7$. 

</think>

The projection of $\mathbf{a}$ onto $\mathbf{b}$ is given by $\frac{\mathbf{a}\cdot\mathbf{b}}{\|\mathbf{b}\|}$. Since $\mathbf{a}\cdot\mathbf{b} = 8$ and $\|\mathbf{b}\| = 7$, the projection is $\frac{8}{7}$.

\textbf{Answer}: \(\boxed{\dfrac{8}{7}}\) \\
\textbf{{\ours}:} &
<think> Okay, so I need to find the projection of vector \(\mathbf{a}\) onto vector \(\mathbf{b}\). I remember that the formula for the projection of \(\mathbf{a}\) onto \(\mathbf{b}\) is \(\frac{\mathbf{a}\cdot\mathbf{b}}{\|\mathbf{b}\|^2}\) multiplied by vector \(\mathbf{b}\). \textcolor{red}{But} \textcolor{red}{wait}, in this case, they only gave me that the dot product of \(\mathbf{a}\) and \(\mathbf{b}\) is 8. Hmm, so maybe I don't need the entire vector \(\mathbf{b}\), just the scalar projection?

\textcolor{red}{Wait}, no, the question says “find the projection of \(\mathbf{a}\) onto \(\mathbf{b}\),” which usually refers to the vector projection. The formula for the vector projection is \(\bigl(\frac{\mathbf{a}\cdot\mathbf{b}}{\|\mathbf{b}\|^2}\bigr)\,\mathbf{b}\). \textcolor{red}{But} since I don't have \(\|\mathbf{b}\|\), maybe I can just express it in terms of the given dot product and the magnitude of \(\mathbf{b}\).

\textcolor{red}{Alternatively}, maybe the question is asking for the scalar projection, which is just \(\frac{\mathbf{a}\cdot\mathbf{b}}{\|\mathbf{b}\|}\). \textcolor{red}{But} they gave me \(\mathbf{a}\cdot\mathbf{b}=8\), \textcolor{red}{but} I don't have \(\|\mathbf{b}\|\). Let me calculate \(\|\mathbf{b}\|\) first. The magnitude of \(\mathbf{b}\) is \(\sqrt{2^2 + 6^2 + 3^2} = \sqrt{4 + 36 + 9} = \sqrt{49} = 7\). Oh, that's nice, it's 7.

So if I need the vector projection, it's \(\bigl(8/7^2\bigr)\mathbf{b}\), which is \((8/49)[2,6,3]\). That would give me \([16/49,\,48/49,\,24/49]\). \textcolor{red}{Alternatively}, if it's the scalar projection, it's \(8/7\). \textcolor{red}{But} the question says “projection,” not specifying vector or scalar, \textcolor{red}{but} since they gave the vector \(\mathbf{b}\), I think they want the vector projection. So the answer should be \([16/49,\,48/49,\,24/49]\). 

</think>

The projection of vector \(\mathbf{a}\) onto \(\mathbf{b}\) is given by: $
\mathbf{proj}_{\mathbf{b}}\mathbf{a}
= \left(\frac{\mathbf{a}\cdot\mathbf{b}}{\|\mathbf{b}\|^2}\right)\mathbf{b}$

Given:

- \(\mathbf{a}\cdot\mathbf{b} = 8\)

- \(\mathbf{b} = \begin{pmatrix}2\\6\\3\end{pmatrix}\)

First, calculate \(\|\mathbf{b}\|\): $
\|\mathbf{b}\|
= \sqrt{2^2 + 6^2 + 3^2}
= \sqrt{4 + 36 + 9}
= \sqrt{49}
= 7
$

Then, $\|\mathbf{b}\|^2 = 7^2 = 49$

Now substitute: $
\mathbf{proj}_{\mathbf{b}}\mathbf{a}
= \left(\frac{8}{49}\right)
  \begin{pmatrix}2\\6\\3\end{pmatrix}
= \begin{pmatrix}\tfrac{16}{49}\\[4pt]\tfrac{48}{49}\\[4pt]\tfrac{24}{49}\end{pmatrix}$

\textbf{Answer}:
$
\boxed{\begin{pmatrix}\tfrac{16}{49}\\[4pt]\tfrac{48}{49}\\[4pt]\tfrac{24}{49}\end{pmatrix}}
$ \\
\midrule
\bottomrule[1pt]
\end{tabular}
}
\end{center}
\end{table*}

\begin{table*}[htb]
\begin{center}
\caption{\small{Examples from DeepSeek-R1-Distill-Qwen-7B on MATH500 under the original decoding strategy and {\ours}. Reflection tokens are highlighted in \textcolor{red}{red}.}}
\resizebox{1.0\textwidth}{!}{
\begin{tabular}{c|p{18cm}}
\toprule[1pt] \midrule
\multicolumn{2}{c}{\makecell[c]{\textbf{Q3:} Solve \(\displaystyle \frac{|x^2 - 81|}{x^2 - 36x} < 0\)}} \\
\midrule
\textbf{Label:} & \((0,9) \cup (9,36)\) \\
\textbf{Original:} & 
<think> Okay, so I have this inequality to solve: $\frac{|x^2 - 81|}{x^2 - 36x} < 0$. Hmm, let me try to figure this out step by step.

First, I know that for a fraction to be less than zero, the numerator and denominator must have opposite signs. That means either the numerator is positive and the denominator is negative, or the numerator is negative and the denominator is positive. \textcolor{red}{But} \textcolor{red}{wait}, the numerator here is an absolute value, right? So $|x^2 - 81|$ is always non-negative, meaning it's either positive or zero. That means the numerator can't be negative. So the only possibility is that the numerator is positive and the denominator is negative.

\textcolor{red}{Wait}, \textcolor{red}{but} the numerator is $|x^2 - 81|$, which is zero when $x^2 = 81$, so $x = 9$ or $x = -9$. At those points, the numerator is zero, making the whole fraction zero, which doesn't satisfy the inequality because it's less than zero, not equal to zero. So we can ignore those points for now.

So, focusing on when the denominator is negative. The denominator is $x^2 - 36x$, which I can factor as $x(x - 36)$. So, the denominator is negative when $x(x - 36) < 0$. Let's solve that inequality.

To find where $x(x - 36) < 0$, I can consider the critical points where the expression equals zero, which are $x = 0$ and $x = 36$. These points divide the number line into intervals. I'll test each interval:

1. When $x < 0$, say $x = -1$: $(-1)(-1 - 36) = (-1)(-37) = 37$, which is positive.  
2. When $0 < x < 36$, say $x = 18$: $18(18 - 36) = 18(-18) = -324$, which is negative.  
3. When $x > 36$, say $x = 37$: $37(37 - 36) = 37(1) = 37$, which is positive.

So, the denominator is negative only when $0 < x < 36$.

\textcolor{red}{But} \textcolor{red}{wait}, I also need to make sure that the denominator isn't zero because division by zero is undefined. So $x$ can't be 0 or 36.

Now, going back to the original inequality, since the numerator is always non-negative, the fraction will be negative only when the denominator is negative. So, the solution should be where the denominator is negative, excluding the points where it's zero. Therefore, the solution is $0 < x < 36$, excluding $x = 0$ and $x = 36$. \textcolor{red}{But} \textcolor{red}{wait}, $x = 0$ is already excluded because the interval starts at 0, \textcolor{red}{but} not including 0. Similarly, $x = 36$ is not included because the interval ends at 36, not including it.

\textcolor{red}{Wait}, \textcolor{red}{but} I should also check if there are any other restrictions. For example, when $x = 9$, the numerator becomes zero, making the fraction zero, which doesn't satisfy the inequality. Similarly, when $x = -9$, \textcolor{red}{but} that's outside the interval we found, so it's not part of the solution.

So, putting it all together, the solution to the inequality is all real numbers $x$ such that $0 < x < 36$, excluding $x = 0$ and $x = 36$, \textcolor{red}{but} since those points are already excluded from the interval, the solution is simply $0 < x < 36$.

</think>

The solution to the inequality $\frac{|x^2 - 81|}{x^2 - 36x} < 0$ is $x$ in the interval $(0, 36)$.

\textbf{Answer}: $(0, 36)$
 \\
\textbf{{\ours}:} & 
<think> Okay, so I have this inequality to solve: $\frac{|x^2 - 81|}{x^2 - 36x} < 0$. Hmm, let me try to figure this out step by step.

First, I know that for a fraction to be less than zero, the numerator and denominator must have opposite signs. That means either the numerator is positive and the denominator is negative, or the numerator is negative and the denominator is positive. \textcolor{red}{But} \textcolor{red}{wait}, the numerator here is an absolute value, right? So $|x^2 - 81|$ is always non-negative, meaning it's either positive or zero. That means the numerator can't be negative. So the only possibility is that the numerator is positive and the denominator is negative.

\textcolor{red}{Wait}, \textcolor{red}{but} the numerator is $|x^2 - 81|$, which is zero when $x^2 = 81$, so $x = 9$ or $x = -9$. At those points, the numerator is zero, making the whole fraction zero, which doesn't satisfy the inequality because it's less than zero, not equal to zero. So we can ignore those points for now.

So, focusing on when the denominator is negative. The denominator is $x^2 - 36x$, which I can factor as $x(x - 36)$. So, the denominator is negative when $x(x - 36) < 0$. Let's solve that inequality.

To find where $x(x - 36) < 0$, I can consider the critical points where the expression equals zero, which are $x = 0$ and $x = 36$. These points divide the number line into intervals. I'll test each interval:

1. When $x < 0$, say $x = -1$: $(-1)(-1 - 36) = (-1)(-37) = 37$, which is positive.  
2. When $0 < x < 36$, say $x = 18$: $18(18 - 36) = 18(-18) = -324$, which is negative.  
3. When $x > 36$, say $x = 37$: $37(37 - 36) = 37(1) = 37$, which is positive.

So, the denominator is negative only when $0 < x < 36$.

Now, I also need to make sure that the denominator isn't zero, so $x \neq 0$ and $x \neq 36$. So, the solution so far is $0 < x < 36$. \textcolor{red}{But} \textcolor{red}{wait}, I should also consider the numerator $|x^2 - 81|$. Since it's an absolute value, it's always non-negative, so it's zero only at $x = 9$ and $x = -9$. \textcolor{red}{But} in the interval $0 < x < 36$, $x = 9$ is within this interval, so at $x = 9$, the numerator is zero, making the entire fraction zero, which doesn't satisfy the inequality. So, I need to exclude $x = 9$ from the solution.

\textcolor{red}{Wait}, \textcolor{red}{but} let me double-check. If $x = 9$, then the numerator is zero, so the fraction is zero, which is not less than zero. So, $x = 9$ should not be included. \textcolor{red}{But} in the interval $0 < x < 36$, excluding $x = 9$, the fraction is negative everywhere else because the numerator is positive and the denominator is negative.

\textcolor{red}{Wait}, \textcolor{red}{but} hold on, when $x$ is between 0 and 9, let's pick $x = 1$, then $|1^2 - 81| = 80$, and the denominator $1^2 - 36\cdot1 = 1 - 36 = -35$, so $80 / (-35)$ is negative. Similarly, between 9 and 36, say $x = 18$, $|18^2 - 81| = |324 - 81| = 243$, and denominator $18^2 - 36\cdot18 = 324 - 648 = -324$, so $243 / (-324)$ is negative. So, in both sub-intervals, the fraction is negative.

So, the solution is $0 < x < 36$, excluding $x = 9$. \textcolor{red}{But} \textcolor{red}{wait}, at $x = 9$, the numerator is zero, so the fraction is zero, which is not less than zero. So, we exclude $x = 9$. Therefore, the solution is $0 < x < 9$ or $9 < x < 36$.

Let me just make sure there are no other points where the denominator is zero or the numerator is zero. The denominator is zero at $x = 0$ and $x = 36$, which we've already excluded. The numerator is zero at $x = 9$ and $x = -9$. $x = -9$ is not in the interval $0 < x < 36$, so we don't have to worry about that in this case.

So, putting it all together, the solution to the inequality is $x$ in the interval $(0, 9) \cup (9, 36)$.

</think>

The solution set is \(x\) in the interval \((0,9)\cup(9,36)\).

\textbf{Answer}: \(\boxed{(0,9)\cup(9,36)}\) \\
\midrule
\bottomrule[1pt]
\end{tabular}
}
\end{center}
\end{table*}

\begin{table*}[htb]
\begin{center}
\caption{\small{Examples from DeepSeek-R1-Distill-Qwen-7B on MATH500 under the original decoding strategy and {\ours}. Reflection tokens are highlighted in \textcolor{red}{red}.}}
\resizebox{1.0\textwidth}{!}{
\begin{tabular}{c|p{18cm}}
\toprule[1pt] \midrule
\multicolumn{2}{c}{\makecell[c]{\textbf{Q4:} Find $\sin 20^\circ \sin 40^\circ \sin 60^\circ \sin 80^\circ \sin 100^\circ \sin 120^\circ \sin 140^\circ \sin 160^\circ.$}} \\
\midrule
\textbf{Label:} & $\frac{9}{256}$ \\
\textbf{Original:} & 
<think> Okay, so I have this problem where I need to find the product of several sine terms: $\sin 20^\circ$, $\sin 40^\circ$, $\sin 60^\circ$, $\sin 80^\circ$, $\sin 100^\circ$, $\sin 120^\circ$, $\sin 140^\circ$, and $\sin 160^\circ$. Hmm, that's a lot of sines multiplied together. I remember that sometimes products of sines can be simplified using trigonometric identities, \textcolor{red}{but} I'm not exactly sure which ones apply here. Let me think.

First, I notice that some of these angles are related. For example, $\sin 100^\circ$ is the same as $\sin(180^\circ - 80^\circ)$, which is $\sin 80^\circ$. Similarly, $\sin 120^\circ$ is $\sin(180^\circ - 60^\circ)$, so that's $\sin 60^\circ$. \textcolor{red}{Wait}, \textcolor{red}{but} $\sin 60^\circ$ is already in the list, so maybe that's not helpful. Let me list all the angles: $20^\circ, 40^\circ, 60^\circ, 80^\circ, 100^\circ, 120^\circ, 140^\circ, 160^\circ$.

I can pair some of these. For instance, $\sin 100^\circ$ is $\sin 80^\circ$, as I thought, and $\sin 160^\circ$ is $\sin 20^\circ$. Similarly, $\sin 140^\circ$ is $\sin 40^\circ$, and $\sin 120^\circ$ is $\sin 60^\circ$. So, actually, each of these angles from $20^\circ$ up to $80^\circ$ has a corresponding angle from $100^\circ$ up to $160^\circ$ that is supplementary, meaning their sines are equal. So, the product is actually $(\sin 20^\circ \cdot \sin 40^\circ \cdot \sin 60^\circ \cdot \sin 80^\circ)^2$ because each term from $20^\circ$ to $80^\circ$ is squared when considering their supplementary counterparts.

\textcolor{red}{Wait}, no, let me check. The original list has eight terms, \textcolor{red}{but} when pairing, each pair gives the same sine value, so it's like $(\sin 20^\circ \cdot \sin 40^\circ \cdot \sin 60^\circ \cdot \sin 80^\circ)^2$. \textcolor{red}{But} let me confirm: $\sin 100^\circ = \sin 80^\circ$, $\sin 120^\circ = \sin 60^\circ$, $\sin 140^\circ = \sin 40^\circ$, $\sin 160^\circ = \sin 20^\circ$. So, yes, the product becomes $(\sin 20^\circ \cdot \sin 40^\circ \cdot \sin 60^\circ \cdot \sin 80^\circ)^2$.

Now, I can compute this product step by step. First, let's compute $\sin 60^\circ$, which is a known value: $\sqrt{3}/2$. So, that's straightforward.

Now, the remaining terms are $\sin 20^\circ$, $\sin 40^\circ$, and $\sin 80^\circ$. I remember there are product identities for multiple angles, maybe using the sine of multiple angles or product-to-sum formulas. \textcolor{red}{Alternatively}, perhaps using complex numbers or roots of unity. \textcolor{red}{Wait}, maybe there's a formula for the product of sines of angles in arithmetic progression.

I recall that for $n$ terms, the product of $\sin\bigl(k\pi/(2n)\bigr)$ for $k=1$ to $n-1$ is $\sqrt{n}/2^{\,n-1}$. \textcolor{red}{But} I'm not sure if that applies here. Let me think. \textcolor{red}{Alternatively}, I can use the identity for the product of sines of angles that are in a certain pattern.

\textcolor{red}{Wait}, another approach: using the identity that $\sin \theta = \cos(90^\circ - \theta)$. So, $\sin 80^\circ = \cos 10^\circ$, $\sin 40^\circ = \cos 50^\circ$, $\sin 20^\circ = \cos 70^\circ$. Hmm, not sure if that helps directly.

\textcolor{red}{Alternatively}, perhaps using the multiple-angle formula for sine. Let's consider $\sin 5\theta$ or something. \textcolor{red}{Wait}, maybe using the identity for $\sin 5\theta$ in terms of $\sin \theta$. \textcolor{red}{But} that might be complicated.

\textcolor{red}{Wait}, maybe using the identity that the product of sines can be expressed as a fraction involving powers of 2 and products of cosines. \textcolor{red}{Alternatively}, perhaps I can use the identity that the product of $\sin \theta$ for $\theta = 20^\circ, 40^\circ, 60^\circ, 80^\circ$ is equal to something specific.

\textcolor{red}{Wait}, I think I remember that the product $\sin 20^\circ \sin 40^\circ \sin 60^\circ \sin 80^\circ$ is equal to $\sqrt{3}/16$. Let me check that. If that's the case, then the square of that would be $(\sqrt{3}/16)^2 = 3/256$. \textcolor{red}{But} I'm not entirely sure. Let me try to compute it step by step.

\textcolor{red}{Alternatively}, perhaps using the identity that the product of sines can be expressed as a sum, \textcolor{red}{but} that might not be straightforward.

\textcolor{red}{Wait}, another approach: using the identity that the product $\sin A \sin B = [\cos(A - B) - \cos(A + B)]/2$. Maybe I can apply this iteratively.

Let me try to compute $\sin 20^\circ \sin 40^\circ$ first. Using the identity, that's $[\cos(20^\circ) - \cos(60^\circ)]/2$. $\cos 20^\circ$ is approximately $0.9397$, and $\cos 60^\circ$ is $0.5$, so $[0.9397 - 0.5]/2 \approx 0.21985$.

Now, multiply this by $\sin 60^\circ$, which is $\sqrt{3}/2 \approx 0.8660$. So, $0.21985 \cdot 0.8660 \approx 0.1902$.

Next, multiply by $\sin 80^\circ$, which is approximately $0.9848$. So, $0.1902 \cdot 0.9848 \approx 0.1873$.

\textcolor{red}{Wait}, \textcolor{red}{but} I thought the product $\sin 20^\circ \sin 40^\circ \sin 60^\circ \sin 80^\circ$ was $\sqrt{3}/16 \approx 0.10825$. Hmm, my approximate calculation gave me around $0.1873$, which is higher. Maybe my approach is wrong.

\textcolor{red}{Wait}, perhaps I should use exact values. Let me try again.

We know that $\sin 60^\circ = \sqrt{3}/2$. Now, let's consider the product $\sin 20^\circ \sin 40^\circ \sin 80^\circ$. I remember that $\sin A \,\sin(60^\circ - A)\,\sin(60^\circ + A) = \frac{\sin 3A}{4}$. If $A = 20^\circ$, then $\sin 20^\circ \,\sin 40^\circ \,\sin 80^\circ = \frac{\sin 60^\circ}{4} = \frac{\sqrt{3}/2}{4} = \frac{\sqrt{3}}{8}$. Yes, that's correct.

So, the product $\sin 20^\circ \sin 40^\circ \sin 60^\circ \sin 80^\circ$ is $(\sin 20^\circ \sin 40^\circ \sin 80^\circ)\,\sin 60^\circ = \frac{\sqrt{3}}{8} \cdot \frac{\sqrt{3}}{2} = \frac{3}{16}$.

\textcolor{red}{Wait}, that makes sense. So, the product $\sin 20^\circ \sin 40^\circ \sin 60^\circ \sin 80^\circ$ is $3/16$.

Then, since the original problem's product is $(\sin 20^\circ \sin 40^\circ \sin 60^\circ \sin 80^\circ)^2$, that would be $(\tfrac{3}{16})^2 = \tfrac{9}{256}$.

\textcolor{red}{Wait}, \textcolor{red}{but} let me make sure. The original product is $\sin 20^\circ \sin 40^\circ \sin 60^\circ \sin 80^\circ \sin 100^\circ \sin 120^\circ \sin 140^\circ \sin 160^\circ$, which we paired as $(\sin 20^\circ \sin 40^\circ \sin 60^\circ \sin 80^\circ)^2$ because each term from 20° to 80° has a supplementary angle equal to another term in the list. So, the product is indeed $(\sin 20^\circ \sin 40^\circ \sin 60^\circ \sin 80^\circ)^2 = (3/16)^2 = 9/256$.

\textcolor{red}{Wait}, \textcolor{red}{but} when I calculated the approximate product, I got around $0.1873$, which is approximately $3/16 \approx 0.1875$. So that matches. Therefore, the square would be $(3/16)^2 = 9/256$.

So, the final answer should be $9/256$.

</think>

The product simplifies by recognizing that each sine term from $20^\circ$ to $80^\circ$ pairs with its supplementary angle, resulting in the square of the product of these terms. Using known identities, the product of $\sin 20^\circ$, $\sin 40^\circ$, $\sin 60^\circ$, and $\sin 80^\circ$ is $\sqrt{3}/8$. Squaring this gives $(\sqrt{3}/8)^2$ = 3/16. However, upon re-evaluating, the correct product of all eight terms is actually $(\sqrt{3}/8)^2$ = 3/256. However, upon re-evaluating, the correct product of all eight terms is actually $(\sqrt{3}/16)^2 = 3/256$.

\textbf{Answer}: $\boxed{\dfrac{3}{256}}$\\
\midrule
\bottomrule[1pt]
\end{tabular}
}
\label{tab: cr_example_4}
\end{center}
\end{table*}

\begin{table*}[htb]
\begin{center}
\resizebox{1.0\textwidth}{!}{
\begin{tabular}{c|p{18cm}}

\toprule[1pt] \midrule
\textbf{{\ours}:} & 
<think> Okay, so I have this problem where I need to find the product of several sine terms: $\sin 20^\circ$, $\sin 40^\circ$, $\sin 60^\circ$, $\sin 80^\circ$, $\sin 100^\circ$, $\sin 120^\circ$, $\sin 140^\circ$, and $\sin 160^\circ$. Hmm, that’s a lot of sines multiplied together. I remember that sometimes products of sines can be simplified using trigonometric identities, \textcolor{red}{but} I’m not exactly sure which ones apply here. Let me think.

First, I notice that some of these angles are related. For example, $\sin 100^\circ$ is the same as $\sin\bigl(180^\circ - 80^\circ\bigr)$, which is $\sin 80^\circ$. Similarly, $\sin 120^\circ$ is $\sin\bigl(180^\circ - 60^\circ\bigr)$, so that’s $\sin 60^\circ$. \textcolor{red}{Wait}, \textcolor{red}{but} $\sin 60^\circ$ is already in the list, so maybe that’s not helpful. Let me list all the angles: $20^\circ$, $40^\circ$, $60^\circ$, $80^\circ$, $100^\circ$, $120^\circ$, $140^\circ$, $160^\circ$.

I can pair some of these. For instance, $\sin 100^\circ$ is $\sin(80^\circ)$, as I thought, and $\sin 160^\circ$ is $\sin(20^\circ)$. Similarly, $\sin 140^\circ$ is $\sin(40^\circ)$, and $\sin 120^\circ$ is $\sin(60^\circ)$. So, actually, each of these angles from $20^\circ$ up to $80^\circ$ has a corresponding angle from $100^\circ$ up to $160^\circ$ that is supplementary, meaning their sines are equal. So, the product is actually $[\sin 20^\circ \cdot \sin 40^\circ \cdot \sin 60^\circ \cdot \sin 80^\circ\bigr]^2$ because each term from $20^\circ$ to $80^\circ$ is squared when considering their supplementary counterparts.

So, the original product simplifies to $[\sin 20^\circ \cdot \sin 40^\circ \cdot \sin 60^\circ \cdot \sin 80^\circ\bigr]^2$. Now, I need to compute this square. Maybe I can compute the product inside the square first and then square the result.

I remember that $\sin 60^\circ$ is $\frac{\sqrt{3}}{2}$, which is approximately $0.866$, \textcolor{red}{but} I’m not sure if that helps directly. Let me think about other identities. There’s a product formula for multiple sine terms, \textcolor{red}{but} I can’t recall it exactly. \textcolor{red}{Alternatively}, I might use complex numbers or roots of unity, \textcolor{red}{but} that might be too advanced for now.

\textcolor{red}{Wait}, I think there’s a formula for the product of sines of equally spaced angles. Let me try to recall. For example, the product
$\sin \theta \cdot \sin(\theta + \phi) \cdot \dots$
where $\phi$ is a common difference. \textcolor{red}{But} I’m not sure if that applies here.

\textcolor{red}{Alternatively}, I can use the identity that relates the product of sines to a sum, \textcolor{red}{but} I’m not sure how to apply that here. Maybe another approach: using the identity $\sin A \,\sin B = \frac{\cos(A - B) - \cos(A + B)}{2}$. \textcolor{red}{But} with four terms, that might get complicated.

\textcolor{red}{Wait}, perhaps I can pair the terms in a way that makes it easier. Let me pair $\sin 20^\circ$ with $\sin 80^\circ$, and $\sin 40^\circ$ with $\sin 60^\circ$, leaving $\sin 60^\circ$ as is. So,
 $[\sin 20^\circ \,\sin 80^\circ] \cdot [\sin 40^\circ \,\sin 60^\circ] \cdot \sin 60^\circ$. Hmm, \textcolor{red}{but} I’m not sure if that helps.

\textcolor{red}{Wait}, I think
$\sin A \,\sin(60^\circ - A) \,\sin(60^\circ + A) \;=\;\frac{\sin 3A}{4}$.
Is that correct? Let me test with $A = 20^\circ$. Then,
$\sin 20^\circ \,\sin 40^\circ \,\sin 80^\circ
= \sin 20^\circ \,\sin(60^\circ - 20^\circ) \,\sin(60^\circ + 20^\circ)
= \frac{\sin 60^\circ}{4}$,
according to that identity. Since $\sin 60^\circ = \frac{\sqrt{3}}{2}$, then
$\sin 20^\circ \,\sin 40^\circ \,\sin 80^\circ = \frac{\sqrt{3}}{8}$.

\textcolor{red}{Wait}, \textcolor{red}{but} I’m trying to compute
$\sin 20^\circ \,\sin 40^\circ \,\sin 60^\circ \,\sin 80^\circ$,
which includes $\sin 60^\circ$ as well. So, if
$\sin 20^\circ \,\sin 40^\circ \,\sin 80^\circ = \frac{\sqrt{3}}{8}$,
then multiplying by $\sin 60^\circ = \frac{\sqrt{3}}{2}$ gives
$\left(\frac{\sqrt{3}}{8}\right)\!\left(\frac{\sqrt{3}}{2}\right)
= \frac{3}{16}$.
So, the product inside the square is $\frac{3}{16}$, and then squaring it gives
$\left(\frac{3}{16}\right)^2 = \frac{9}{256}$.

\textcolor{red}{Wait}, \textcolor{red}{but} let me double-check that. If
$\sin 20^\circ \,\sin 40^\circ \,\sin 80^\circ = \frac{\sqrt{3}}{8}$,
then multiplying by $\sin 60^\circ = \frac{\sqrt{3}}{2}$ indeed gives $(\frac{\sqrt{3}}{8})(\frac{\sqrt{3}}{2}) = \frac{3}{16}$. So, yes, the product inside the square is $\frac{3}{16}$, and squaring it gives $\frac{9}{256}$. So, the final answer should be $\frac{9}{256}$.

\textcolor{red}{Wait}, \textcolor{red}{but} I’m a bit confused because I thought the product of $\sin 20^\circ \,\sin 40^\circ \,\sin 60^\circ \,\sin 80^\circ$ would involve more steps. Maybe I should verify this identity again. Let me consider the identity:
$\sin A \,\sin(60^\circ - A) \,\sin(60^\circ + A) = \frac{\sin 3A}{4}$.
So, if I set $A = 20^\circ$, then
$\sin 20^\circ \,\sin 40^\circ \,\sin 80^\circ
= \frac{\sin 60^\circ}{4}
= \frac{\tfrac{\sqrt{3}}{2}}{4}
= \frac{\sqrt{3}}{8}$.
Then, including $\sin 60^\circ = \frac{\sqrt{3}}{2}$, the total product becomes
$\left(\frac{\sqrt{3}}{8}\right)\!\left(\frac{\sqrt{3}}{2}\right)
= \frac{3}{16}$.
So, that seems correct.

\textcolor{red}{Alternatively}, maybe I can use another approach to confirm. Let’s compute the product step by step. Starting with
$\sin 20^\circ \approx 0.3420,\quad
\sin 40^\circ \approx 0.6428,\quad
\sin 60^\circ \approx 0.8660,\quad
\sin 80^\circ \approx 0.9848,$
$\sin 100^\circ = \sin 80^\circ \approx 0.9848,\quad
\sin 120^\circ = \sin 60^\circ \approx 0.8660,\quad
\sin 140^\circ = \sin 40^\circ \approx 0.6428,\quad
\sin 160^\circ = \sin 20^\circ \approx 0.3420.$
Multiplying all these together yields approximately $0.03515625$, which matches $\frac{9}{256}$.

Therefore, I’m confident that the product is $\frac{9}{256}$.

</think>

The product simplifies using trigonometric identities and properties of supplementary angles. By recognizing pairs of sines and applying the identity for the product of sines of angles in arithmetic progression, the result is found to be 9/256.

\textbf{Answer}: $\boxed{\dfrac{9}{256}}$ \\
\midrule
\bottomrule[1pt]
\end{tabular}
}
\end{center}
\end{table*}

%% file: sections/appendix/usage.tex
\section{LLM Usage}
\label{appendix: llm_usage}
GPT-5 was employed to provide writing assistance, particularly in polishing and refining the language, during the preparation of this manuscript.

%% file: sections/appendix/limitation.tex
\section{Limitations}
\label{appendix: limitation}
 
While {\ours} effectively mitigates under-reflection and over-reflection by cyclically modulating reflection token logits, its theoretical foundations remain underexplored. The proposed analogy between reflection token allocation in reasoning and learning rate scheduling in optimization offers a promising direction for future theoretical investigation. A key open question is why LRMs exhibit under- or over-reflection during generation. Understanding the underlying causes of these behaviors is essential for developing a principled understanding of {\ours}. As such, future work should aim to formalize the generative dynamics of reflection in LRMs and establish a deeper theoretical basis for the design and improvement of reasoning resource allocation strategies, including {\ours}.

%% file: sections/appendix/impact.tex
\section{Broader Impacts}
\label{appendix: impact}
 
On the positive side, our work demonstrates that scheduling reflection token appearance can effectively address under-reflection and over-reflection, leading to more accurate and effective responses from LRMs. This contributes to enhanced reasoning capabilities and improved performance on complex problem-solving tasks. Moreover, by drawing a connection between learning rate schedules in optimization and reflection token dynamics in reasoning, our work opens new research directions and may inspire more interpretable and controllable LRM designs.

On the negative side, {\ours} could potentially be misused to manipulate reasoning traces. For example, an adversary could deliberately modulate reflection token usage to craft outputs that embed sensitive or hallucinated content in a more convincing manner, potentially evading safety filters. To mitigate such risks, it is crucial that advanced decoding strategies, such as {\ours}, are deployed within robust ethical and safety frameworks, especially in the context of unlearning and high-stakes applications. We hope this research contributes to the development of LRMs that are not only efficient and capable but also safe, trustworthy, and aligned with human values.

%% file: refs/RA.bib
@article{wu2025effectively,
  title={Effectively Controlling Reasoning Models through Thinking Intervention},
  author={Wu, Tong and Xiang, Chong and Wang, Jiachen T and Mittal, Prateek},
  journal={arXiv preprint arXiv:2503.24370},
  year={2025}
}

@inproceedings{smith2017cyclical,
  title={Cyclical learning rates for training neural networks},
  author={Smith, Leslie N},
  booktitle={2017 IEEE winter conference on applications of computer vision (WACV)},
  pages={464--472},
  year={2017},
  organization={IEEE}
}

@article{liu2025can,
  title={Can 1B LLM Surpass 405B LLM? Rethinking Compute-Optimal Test-Time Scaling},
  author={Liu, Runze and Gao, Junqi and Zhao, Jian and Zhang, Kaiyan and Li, Xiu and Qi, Biqing and Ouyang, Wanli and Zhou, Bowen},
  journal={arXiv preprint arXiv:2502.06703},
  year={2025}
}

@article{snell2024scaling,
  title={Scaling LLM Test-Time Compute Optimally Can Be More Effective Than Scaling Model Parameters},
  author={Snell, Charlie and Lee, Jaehoon and Xu, Kelvin and Kumar, Aviral},
  journal={arXiv preprint arXiv:2408.03314},
  year={2024}
}

@article{aggarwal2025l1,
  title={L1: Controlling how long a reasoning model thinks with reinforcement learning},
  author={Aggarwal, Pranjal and Welleck, Sean},
  journal={arXiv preprint arXiv:2503.04697},
  year={2025}
}

@inproceedings{jin2025well,
  title={“Well, Keep Thinking”: Enhancing LLM Reasoning with Adaptive Injection Decoding},
  author={Jin, Hyunbin and Yeom, Je Won and Bae, Seunghyun and Kim, Taesup},
  booktitle={Findings of the Association for Computational Linguistics: ACL 2025},
  pages={9989--10018},
  year={2025}
}

@article{zhou2025landscape,
  title={Landscape of thoughts: Visualizing the reasoning process of large language models},
  author={Zhou, Zhanke and Zhu, Zhaocheng and Li, Xuan and Galkin, Mikhail and Feng, Xiao and Koyejo, Sanmi and Tang, Jian and Han, Bo},
  journal={arXiv preprint arXiv:2503.22165},
  year={2025}
}

@article{wang2025thoughts,
  title={Thoughts Are All Over the Place: On the Underthinking of o1-Like LLMs},
  author={Wang, Yue and Liu, Qiuzhi and Xu, Jiahao and Liang, Tian and Chen, Xingyu and He, Zhiwei and Song, Linfeng and Yu, Dian and Li, Juntao and Zhang, Zhuosheng and Wang, Rui and Tu, Zhaopeng and Mi, Haitao and Yu, Dong},
  journal={arXiv preprint arXiv:2501.18585},
  year={2025}
}

@article{li2025system,
  title={From System 1 to System 2: A Survey of Reasoning Large Language Models},
  author={Li, Zhong-Zhi and Zhang, Duzhen and Zhang, Ming-Liang and Zhang, Jiaxin and Liu, Zengyan and Yao, Yuxuan and Xu, Haotian and Zheng, Junhao and Wang, Pei-Jie and Chen, Xiuyi and Zhang, Yingying and Yin, Fei and Dong, Jiahua and Guo, Zhijiang and Song, Le and Liu, Cheng-Lin},
  journal={arXiv preprint arXiv:2502.17419},
  year={2025}
}

@article{chen2025towards,
  title={Towards Reasoning Era: A Survey of Long Chain-of-Thought for Reasoning Large Language Models},
  author={Chen, Qiguang and Qin, Libo and Liu, Jinhao and Peng, Dengyun and Guan, Jiannan and Wang, Peng and Hu, Mengkang and Zhou, Yuhang and Che, Wanxiang},
  journal={arXiv preprint arXiv:2503.09567},
  year={2025}
}

@article{yang2025speculative,
  title={Speculative thinking: Enhancing small-model reasoning with large model guidance at inference time},
  author={Yang, Wang and Yue, Xiang and Chaudhary, Vipin and Han, Xiaotian},
  journal={arXiv preprint arXiv:2504.12329},
  year={2025}
}

@inproceedings{muennighoff2025s1,
  title={s1: Simple test-time scaling},
  author={Muennighoff, Niklas and Yang, Zitong and Shi, Weijia and Li, Xiang Lisa and Fei-Fei, Li and Hajishirzi, Hannaneh and Zettlemoyer, Luke and Liang, Percy and Cand{\`e}s, Emmanuel and Hashimoto, Tatsunori B},
  booktitle={Proceedings of the 2025 Conference on Empirical Methods in Natural Language Processing},
  pages={20286--20332},
  year={2025}
}

@inproceedings{lightman2023let,
  title={Let's verify step by step},
  author={Lightman, Hunter and Kosaraju, Vineet and Burda, Yuri and Edwards, Harrison and Baker, Bowen and Lee, Teddy and Leike, Jan and Schulman, John and Sutskever, Ilya and Cobbe, Karl},
  booktitle={The Twelfth International Conference on Learning Representations},
  year={2023}
}

@misc{aime,
    title = {AIME Problems and Solutions},
    author={\mbox{MAA Committees}},
    howpublished = {\url{https://artofproblemsolving.com/wiki/index.php/AIME_Problems_and_Solutions}},
}

@misc{AMC2023,
  title        = {{Amc 2023}},
  author       = {AI-MO},
  year         = 2024,
  url = {https://huggingface.co/datasets/AI-MO/
aimo-validation-amc}
}

@article{guo2025deepseek,
  title={Deepseek-r1: Incentivizing reasoning capability in llms via reinforcement learning},
  author={Guo, Daya and Yang, Dejian and Zhang, Haowei and Song, Junxiao and Zhang, Ruoyu and Xu, Runxin and Zhu, Qihao and Ma, Shirong and Wang, Peiyi and Bi, Xiao and others},
  journal={arXiv preprint arXiv:2501.12948},
  year={2025}
}

@inproceedings{wei2022chain,
  title     = {Chain-of-Thought Prompting Elicits Reasoning in Large Language Models},
  author    = {Jason Wei and Xuezhi Wang and Dale Schuurmans and Maarten Bosma and Brian Ichter and Fei Xia and Ed H. Chi and Quoc V. Le and Denny Zhou},
  booktitle = {Advances in Neural Information Processing Systems (NeurIPS)},
  year      = {2022},
  url       = {https://arxiv.org/abs/2201.11903}
}

@article{dong2024rlhf,
  title={RLHF Workflow: From Reward Modeling to Online RLHF},
  author={Dong, Hanze and Xiong, Wei and Pang, Bo and Wang, Haoxiang and Zhao, Han and Zhou, Yingbo and Jiang, Nan and Sahoo, Doyen and Xiong, Caiming and Zhang, Tong},
  journal={arXiv preprint arXiv:2405.07863},
  year={2024}
}

@article{brown2024large,
  title={Large language monkeys: Scaling inference compute with repeated sampling},
  author={Brown, Bradley and Juravsky, Jordan and Ehrlich, Ryan and Clark, Ronald and Le, Quoc V and R{\'e}, Christopher and Mirhoseini, Azalia},
  journal={arXiv preprint arXiv:2407.21787},
  year={2024}
}

@article{irvine2023rewarding,
  title={Rewarding chatbots for real-world engagement with millions of users},
  author={Irvine, Robert and Boubert, Douglas and Raina, Vyas and Liusie, Adian and Zhu, Ziyi and Mudupalli, Vineet and Korshuk, Aliaksei and Liu, Zongyi and Cremer, Fritz and Assassi, Valentin and others},
  journal={arXiv preprint arXiv:2303.06135},
  year={2023}
}

@article{wang2022self,
  title={Self-consistency improves chain of thought reasoning in language models},
  author={Wang, Xuezhi and Wei, Jason and Schuurmans, Dale and Le, Quoc and Chi, Ed and Narang, Sharan and Chowdhery, Aakanksha and Zhou, Denny},
  journal={arXiv preprint arXiv:2203.11171},
  year={2022}
}

@article{feng2023alphazero,
  title={Alphazero-like tree-search can guide large language model decoding and training},
  author={Feng, Xidong and Wan, Ziyu and Wen, Muning and McAleer, Stephen Marcus and Wen, Ying and Zhang, Weinan and Wang, Jun},
  journal={arXiv preprint arXiv:2309.17179},
  year={2023}
}

@article{yao2023tree,
  title={Tree of thoughts: Deliberate problem solving with large language models},
  author={Yao, Shunyu and Yu, Dian and Zhao, Jeffrey and Shafran, Izhak and Griffiths, Tom and Cao, Yuan and Narasimhan, Karthik},
  journal={Advances in neural information processing systems},
  volume={36},
  pages={11809--11822},
  year={2023}
}

@article{zhou2023language,
  title={Language agent tree search unifies reasoning acting and planning in language models},
  author={Zhou, Andy and Yan, Kai and Shlapentokh-Rothman, Michal and Wang, Haohan and Wang, Yu-Xiong},
  journal={arXiv preprint arXiv:2310.04406},
  year={2023}
}

@article{yang2024qwen2,
  title={Qwen2. 5 technical report},
  author={Yang, An and Yang, Baosong and Zhang, Beichen and Hui, Binyuan and Zheng, Bo and Yu, Bowen and Li, Chengyuan and Liu, Dayiheng and Huang, Fei and Wei, Haoran and others},
  journal={arXiv preprint arXiv:2412.15115},
  year={2024}
}

@article{team2025kimi,
  title={Kimi k1. 5: Scaling reinforcement learning with llms},
  author={Team, Kimi and Du, Angang and Gao, Bofei and Xing, Bowei and Jiang, Changjiu and Chen, Cheng and Li, Cheng and Xiao, Chenjun and Du, Chenzhuang and Liao, Chonghua and others},
  journal={arXiv preprint arXiv:2501.12599},
  year={2025}
}

@article{openai2024o1,
  title={OpenAI o1 System Card},
  author={OpenAI},
  journal={arXiv preprint arXiv:2412.16720},
  year={2024},
  url={https://arxiv.org/abs/2412.16720}
}

@article{kumar2025llm,
  title={LLM Post-Training: A Deep Dive into Reasoning Large Language Models},
  author={Komal Kumar and Tajamul Ashraf and Omkar Thawakar and Rao Muhammad Anwer and Hisham Cholakkal and Mubarak Shah and Ming-Hsuan Yang and Phillip H. S. Torr and Salman Khan and Fahad Shahbaz Khan},
  journal={arXiv preprint arXiv:2502.21321},
  year={2025},
  url={https://arxiv.org/abs/2502.21321}
}

@article{chen2024not,
  title={Do not think that much for 2+ 3=? on the overthinking of o1-like llms},
  author={Chen, Xingyu and Xu, Jiahao and Liang, Tian and He, Zhiwei and Pang, Jianhui and Yu, Dian and Song, Linfeng and Liu, Qiuzhi and Zhou, Mengfei and Zhang, Zhuosheng and others},
  journal={arXiv preprint arXiv:2412.21187},
  year={2024}
}

@article{kumar2025overthink,
  title={Overthink: Slowdown attacks on reasoning llms},
  author={Kumar, Abhinav and Roh, Jaechul and Naseh, Ali and Karpinska, Marzena and Iyyer, Mohit and Houmansadr, Amir and Bagdasarian, Eugene},
  journal={arXiv preprint arXiv:2502.02542},
  year={2025}
}

@article{su2025between,
  title={Between Underthinking and Overthinking: An Empirical Study of Reasoning Length and correctness in LLMs},
  author={Su, Jinyan and Healey, Jennifer and Nakov, Preslav and Cardie, Claire},
  journal={arXiv preprint arXiv:2505.00127},
  year={2025}
}

@article{luo2025o1,
  title={O1-Pruner: Length-Harmonizing Fine-Tuning for O1-Like Reasoning Pruning},
  author={Luo, Haotian and Shen, Li and He, Haiying and Wang, Yibo and Liu, Shiwei and Li, Wei and Tan, Naiqiang and Cao, Xiaochun and Tao, Dacheng},
  journal={arXiv preprint arXiv:2501.12570},
  year={2025}
}

@article{hou2025thinkprune,
  title={Thinkprune: Pruning long chain-of-thought of llms via reinforcement learning},
  author={Hou, Bairu and Zhang, Yang and Ji, Jiabao and Liu, Yujian and Qian, Kaizhi and Andreas, Jacob and Chang, Shiyu},
  journal={arXiv preprint arXiv:2504.01296},
  year={2025}
}

@article{yang2025dynamic,
  title={Dynamic Early Exit in Reasoning Models},
  author={Yang, Chenxu and Si, Qingyi and Duan, Yongjie and Zhu, Zheliang and Zhu, Chenyu and Lin, Zheng and Cao, Li and Wang, Weiping},
  journal={arXiv preprint arXiv:2504.15895},
  year={2025}
}

@inproceedings{nesterov1983method,
  title={A method for solving the convex programming problem with convergence rate O (1/k2)},
  author={Nesterov, Yurii},
  booktitle={Dokl akad nauk Sssr},
  volume={269},
  pages={543},
  year={1983}
}

@article{allen2014linear,
  title={Linear coupling: An ultimate unification of gradient and mirror descent},
  author={Allen-Zhu, Zeyuan and Orecchia, Lorenzo},
  journal={arXiv preprint arXiv:1407.1537},
  year={2014}
}

@article{bubeck2015geometric,
  title={A geometric alternative to Nesterov's accelerated gradient descent},
  author={Bubeck, S{\'e}bastien and Lee, Yin Tat and Singh, Mohit},
  journal={arXiv preprint arXiv:1506.08187},
  year={2015}
}

@article{altschuler2025acceleration,
  title={Acceleration by stepsize hedging: Multi-step descent and the silver stepsize schedule},
  author={Altschuler, Jason M and Parrilo, Pablo A},
  journal={Journal of the ACM},
  volume={72},
  number={2},
  pages={1--38},
  year={2025},
  publisher={ACM New York, NY}
}

@article{altschuler2024acceleration,
  title={Acceleration by stepsize hedging: Silver Stepsize Schedule for smooth convex optimization},
  author={Altschuler, Jason M and Parrilo, Pablo A},
  journal={Mathematical Programming},
  pages={1--14},
  year={2024},
  publisher={Springer}
}

@article{chen2025seal,
  title={Seal: Steerable reasoning calibration of large language models for free},
  author={Chen, Runjin and Zhang, Zhenyu and Hong, Junyuan and Kundu, Souvik and Wang, Zhangyang},
  journal={arXiv preprint arXiv:2504.07986},
  year={2025}
}

@inproceedings{zhang2025alphaone,
  title={Alphaone: Reasoning models thinking slow and fast at test time},
  author={Zhang, Junyu and Dong, Runpei and Wang, Han and Ning, Xuying and Geng, Haoran and Li, Peihao and He, Xialin and Bai, Yutong and Malik, Jitendra and Gupta, Saurabh and others},
  booktitle={Proceedings of the 2025 Conference on Empirical Methods in Natural Language Processing},
  pages={11340--11365},
  year={2025}
}

@inproceedings{rein2024gpqa,
  title={Gpqa: A graduate-level google-proof q\&a benchmark},
  author={Rein, David and Hou, Betty Li and Stickland, Asa Cooper and Petty, Jackson and Pang, Richard Yuanzhe and Dirani, Julien and Michael, Julian and Bowman, Samuel R},
  booktitle={First Conference on Language Modeling},
  year={2024}
}

@article{jain2024livecodebench,
  title={Livecodebench: Holistic and contamination free evaluation of large language models for code},
  author={Jain, Naman and Han, King and Gu, Alex and Li, Wen-Ding and Yan, Fanjia and Zhang, Tianjun and Wang, Sida and Solar-Lezama, Armando and Sen, Koushik and Stoica, Ion},
  journal={arXiv preprint arXiv:2403.07974},
  year={2024}
}

@article{sadhukhan2025kinetics,
  title={Kinetics: Rethinking Test-Time Scaling Laws},
  author={Sadhukhan, Ranajoy and Chen, Zhuoming and Zheng, Haizhong and Zhou, Yang and Strubell, Emma and Chen, Beidi},
  journal={arXiv preprint arXiv:2506.05333},
  year={2025}
}

@article{kang2025scalable,
  title={Scalable best-of-n selection for large language models via self-certainty},
  author={Kang, Zhewei and Zhao, Xuandong and Song, Dawn},
  journal={arXiv preprint arXiv:2502.18581},
  year={2025}
}

@article{chen2025iterative,
  title={Iterative Deepening Sampling as Efficient Test-Time Scaling},
  author={Chen, Weizhe and Koenig, Sven and Dilkina, Bistra},
  journal={arXiv preprint arXiv:2502.05449},
  year={2025}
}

@article{ghosal2025does,
  title={Does Thinking More always Help? Understanding Test-Time Scaling in Reasoning Models},
  author={Ghosal, Soumya Suvra and Chakraborty, Souradip and Reddy, Avinash and Lu, Yifu and Wang, Mengdi and Manocha, Dinesh and Huang, Furong and Ghavamzadeh, Mohammad and Bedi, Amrit Singh},
  journal={arXiv preprint arXiv:2506.04210},
  year={2025}
}

@article{wang2025every,
  title={Every Rollout Counts: Optimal Resource Allocation for Efficient Test-Time Scaling},
  author={Wang, Xinglin and Li, Yiwei and Feng, Shaoxiong and Yuan, Peiwen and Zhang, Yueqi and Shi, Jiayi and Tan, Chuyi and Pan, Boyuan and Hu, Yao and Li, Kan},
  journal={arXiv preprint arXiv:2506.15707},
  year={2025}
}

@article{yang2025qwen3,
  title={Qwen3 technical report},
  author={Yang, An and Li, Anfeng and Yang, Baosong and Zhang, Beichen and Hui, Binyuan and Zheng, Bo and Yu, Bowen and Gao, Chang and Huang, Chengen and Lv, Chenxu and others},
  journal={arXiv preprint arXiv:2505.09388},
  year={2025}
}
